\documentclass[5p,times]{elsarticle}
\usepackage{framed,multirow}
\usepackage{empheq}
\usepackage{amssymb}
\usepackage{latexsym}

\usepackage{url}
\usepackage{xcolor}

\usepackage{hyperref}
\usepackage{enumitem}

\usepackage{lineno}
\usepackage{verbatim}
\usepackage{nicefrac}
\usepackage{amsmath}
\usepackage{hyperref}
\usepackage[algo2e,ruled,vlined]{algorithm2e}
\usepackage{bbm}
\usepackage{upgreek}
\usepackage{amssymb, graphicx, booktabs, xcolor, upgreek, algorithm, algpseudocode, changepage}

\definecolor{newcolor}{rgb}{.8,.349,.1}

\definecolor{mygreen}{rgb}{0.0431,0.4000,0.1373}

\definecolor{myred}{rgb}{0.89, 0.0, 0.13}

\usepackage{todonotes}

\newcommand{\benoit}[1]{\todo[fancyline,backgroundcolor=yellow!25,bordercolor=yellow]{\footnotesize{}Benoit says: #1}}

\newcommand{\rred}[1]{\textcolor{red}{#1}}

\newcommand{\figNorm}[1]{
\includegraphics[height=.4\linewidth, width=.33\linewidth]{figures/normalization/#1.pdf}}

\newcommand{\params}{\uptheta}
\newcommand{\expect}{\mathbb{E}}

\newcommand{\B}{\bfseries}

\renewcommand{\vec}[1]{\mathbf{#1}}
\newcommand{\mr}[1]{\mathrm{#1}}

\newcommand{\yy}{\vec{y}}
\newcommand{\xx}{\vec{x}}

\newcommand{\sss}{\vec{s}}
\newcommand{\xnorm}{\widehat{\vec{x}}}

\newcommand{\data}{\mathcal{D}}
\newcommand{\real}{\mathbb{R}}
\newcommand{\loss}{\mathcal{L}}
\newcommand{\lossSeg}{\loss_{\mr{seg}}}
\newcommand{\lossDis}{\loss_{\mr{dis}}}
\newcommand{\img}{\Upomega}
\newcommand{\ttm}{$\times$}

\newcommand{\mypar}[1]{\paragraph{\bf{}#1}}

\newcommand{\confmat}[1]{
 \includegraphics[width=.24\linewidth,height=.26\linewidth]{figures/confusion_mat/#1.pdf}}
 
\newtheorem{thm}{Theorem}

\newdefinition{rmk}{Remark}
\newproof{pf}{Proof}
\newproof{pot}{Proof of Theorem \ref{thm2}}

\journal{Medical Image Analysis}

\begin{document}

% \verso{Delisle \textit{et~al.}}
\begin{frontmatter}

\title{Realistic Image Normalization for Multi-Domain Segmentation}%

%\tnotetext[tnote1]{This is an example for title footnote coding.}

\author[1]{Pierre-Luc Delisle\corref{cor1}}
\cortext[cor1]{Corresponding author: 
  pierre-luc.delisle.1@etsmtl.net}
\author[1]{Benoit Anctil-Robitaille}
%\fntext[fn1]{This is author footnote for second author.}
\author[1]{Christian Desrosiers}
%% Third author's email
%\ead{author3@author.com}
\author[1]{Herve Lombaert}

\address[1]{Department of Computer and Software Engineering, ETS Montreal, Canada}

\begin{abstract}
%%%
Image normalization is a building block in medical image analysis. Conventional approaches are customarily utilized on a per-dataset basis. This strategy, however, prevents the current normalization algorithms from fully exploiting the complex joint information available across multiple datasets. Consequently, ignoring such joint information has a direct impact on the performance of segmentation algorithms. This paper proposes to revisit the conventional image normalization approach by instead learning a common normalizing function across multiple datasets. Jointly normalizing multiple datasets is shown to yield consistent normalized images as well as an improved image segmentation. To do so, a fully automated adversarial and task-driven normalization approach is employed as it facilitates the training of realistic and interpretable images while keeping performance on-par with the state-of-the-art. The adversarial training of our network aims at finding the optimal transfer function to improve both the segmentation accuracy and the generation of realistic images. We evaluated the performance of our normalizer on both infant and adult brains images from the iSEG, MRBrainS and ABIDE datasets. Results reveal the potential of our normalization approach for segmentation, with Dice improvements of up to 57.5\% over our baseline. Our method can also enhance data availability by increasing the number of samples available when learning from multiple imaging domains.
%%%%
\end{abstract}


\end{frontmatter}

%\linenumbers

\section{Introduction}
\label{S:1}
Powered by their capacity to learn hierarchical feature representations from data, deep learning algorithms have achieved unprecedented performance in a broad range of medical imaging applications. Notably, deep convolutional neural networks (CNNs) have helped improve the segmentation of various anatomical structures in medical images, for instance brain regions in 3D MRI~\citep{Dolz2019, Kamnitsas2017}. However, supervised learning algorithms typically require a large amount of labeled data for training. Obtaining such quantities is often difficult, since the manual labeling of images is a complex and time-consuming process performed by highly-trained clinical experts. 

A possible approach to both alleviate the lack of training data and increase the generalization performance of the learning algorithm is to use data acquired from multiple sites. However, medical images from separate datasets can be acquired with distinct scanner models or parameters, and therefore may present drastic differences in their images intensities. Acquisition standards could be normalized but reconstructed images still exhibit important differences across sites ~\citep{Kochunov2014}. Another potential source of variability in intensities can also arise from differences in patient demographics between datasets. This problem is illustrated in Fig.~\ref{figure1}, where image intensity histograms are shown for two public datasets, MRBrainS, on adult brains, and iSEG, on 6-8 months old infants. These histograms indicate differences in the distribution overlap of tissue classes across datasets, which directly impair any subsequent segmentation processing.

Standard deep learning models are sensitive to the data distribution on which they are trained. This leads to sub-optimal performance when evaluating on different sets of medical images~\citep{maartensson2020reliability}. A common strategy to address this problem is to normalize images in a pre-processing step, for instance, so that their intensities fall in the same range or have the same global mean~\citep{Onofrey2019}. However, this naive approach is generally insufficient for tasks such as segmentation since it does not consider the intensity distribution of individual regions in the image. Hence, a normalization of images yielding the same global average intensity may still result in different class-specific distributions, directly impacting any downstream analysis.

\begin{figure}[t!]
\centering

    \includegraphics[width=\linewidth]{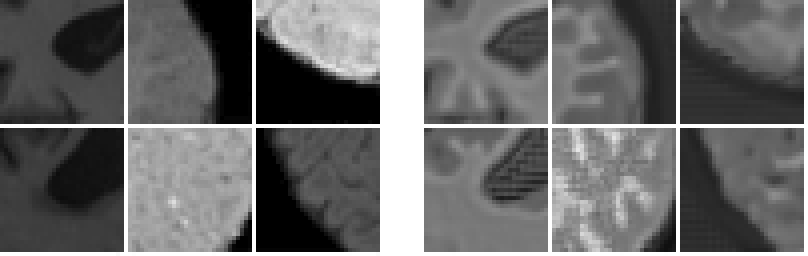}
    
    \vspace*{-2mm}
    \caption{Mixed iSEG and MRBrainS inputs (\textbf{left}) and images generated with two pipelined FCNs without constraint on realism using only Dice loss (\textbf{right}). Images generated with only Dice loss preserve the structure required for segmentation but lack realism.}
    \label{fig:realism}
\end{figure}

Recent work has investigated the potential of deep neural networks for data-driven image normalization. Instead of feeding images directly to a segmentation network, \cite{Drozdzal2018} employed a second CNN as a pre-processing network to normalize images prior to their segmentation. Their strategy of learning image-specific normalization has led to a better segmentation performance with images of different characteristics. However, as shown in Fig. \ref{fig:realism}, since there is no constraint on the realism of images produced by the normalization network, these images typically lack interpretability across datasets. While their work has focused on a single data source, other studies have considered the problem of harmonizing data across multiple sites. For instance, DeepHarmony~\citep{dewey2019deepharmony} uses a fully-convolutional CNN architecture to translate images from one acquisition protocol to another. Despite improved results, important limitations remain:  1) it requires having images of the same subjects for different protocols; 2) it cannot be easily extended to more than two sites since it relies on learning a protocol-to-protocol mapping; 3) it still needs several pre-processing steps to mitigate image inhomogeneity and perform gain correction. A normalizing approach without such paired images of the same subjects across protocols has been explored~\citep{modanwal2020mri} using a cycle-consistent generative adversarial network (CycleGAN). However, such approach is not tailored to a specific task such as segmentation, and can consequently lead to sub-optimal results in the downstream analyses. Nonetheless, GANs proved to be powerful at generating medical images \citep{Nie2020}.

\subsection{Contributions}

We address the limitations of existing image normalization approaches with a novel adversarial learning method that generates normalized images that are both interpretable by clinicians and optimized for a downstream segmentation task. Our method leverages information from multiple datasets by learning a joint normalizing transformation accounting for large image variability. This is achieved with a deep learning architecture comprised of two fully-convolutional 3D CNNs~\citep{Long_2015_CVPR}, the first one acting as a normalized image generator and the second one used as segmentation network. During training, our model also includes a 3D CNN-based discriminator~\citep{RN12} which serves as a domain classifier. In standard adversarial learning approaches for image generation, the discriminator tries to classify images as real or fake~\citep{Goodfellow2014}. On the other hand, typical approaches for domain adaptation instead use the discriminator for predicting if an image is from a source or a target domain~\citep{Kamnitsas2017,vanOpbroek2015Transfer,Cheplygina2018Notsosupervised}. In our proposed normalization method, the discriminator distinguishes images between all input domains (i.e., acquisition site and/or protocol) \emph{as well} as an additional ``generated'' class. Hence, the produced images are both realistic and domain-invariant.

Our contributions can be summarized as follows: 
\begin{itemize}[itemsep=1pt,topsep=2pt]
    \item A first learned normalization method for medical images producing images that are both optimized for segmentation and interpretable by clinicians. Compared to recent approaches, such as those based on CycleGANs, our method can accommodate an arbitrary number of data source domains without additional complexity.   

    \item A novel adversarial learning model for 3D image processing that jointly optimizes three convolutional neural networks. Unlike standard adversarial techniques which have separate discriminators for domain classification and differentiating generated images from real ones, our model combines these two tasks in a single network using an additional domain class. As theoretical contribution, we show that optimizing this model corresponds to minimizing the KL divergence between the generated ima\-ge probability distribution of each domain and the mean distribution of real images.

    \item One of the most comprehensive experimental analyses of learned normalization models for medical images, which includes three very dif\-fe\-rent brain MRI datasets and also evaluates our method on multi-modal data and images degraded by bias field. This analysis demonstrates the advantages of our approach compared to the recent state-of-the-art.
    
\end{itemize}

After an overview of related work, the methodology describes our task-driven normalization method. The experimental results follow with an evaluation on: 1) the cross-data performance of a standard supervised baseline; 2) the normalization and segmentation performance of our method on data from two or more sites; 3) the capacity to normalize multi-modal images; and 4) the ability to correct for intensity inhomogeneity in images with strong bias field. The conclusion finally summarizes the main contributions and results of our work, and discusses possible extensions. 

\begin{figure}[t!]
    \centering
      \includegraphics[width=.8\linewidth]{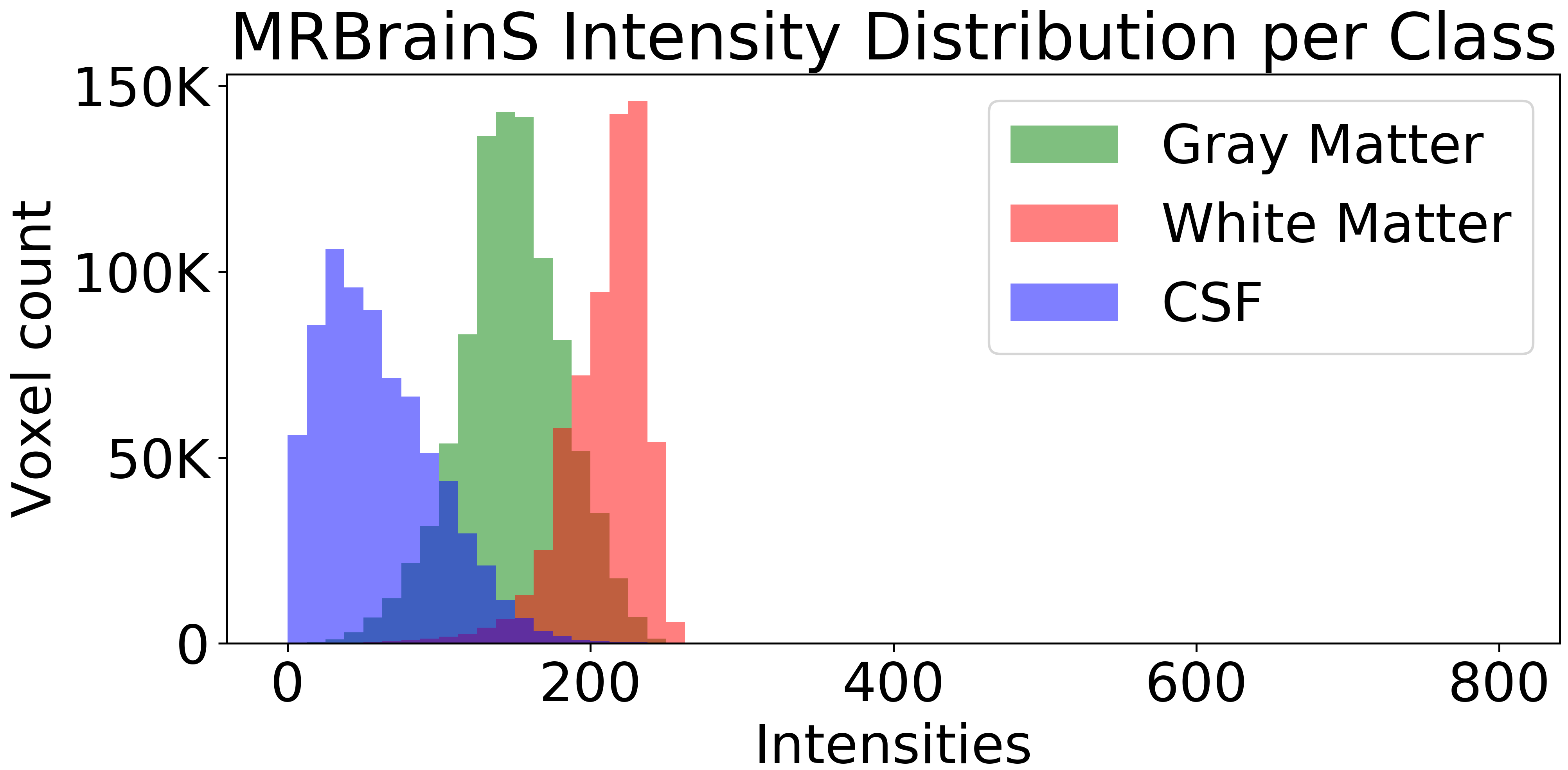} \\[8pt]
      \includegraphics[width=.8\linewidth]{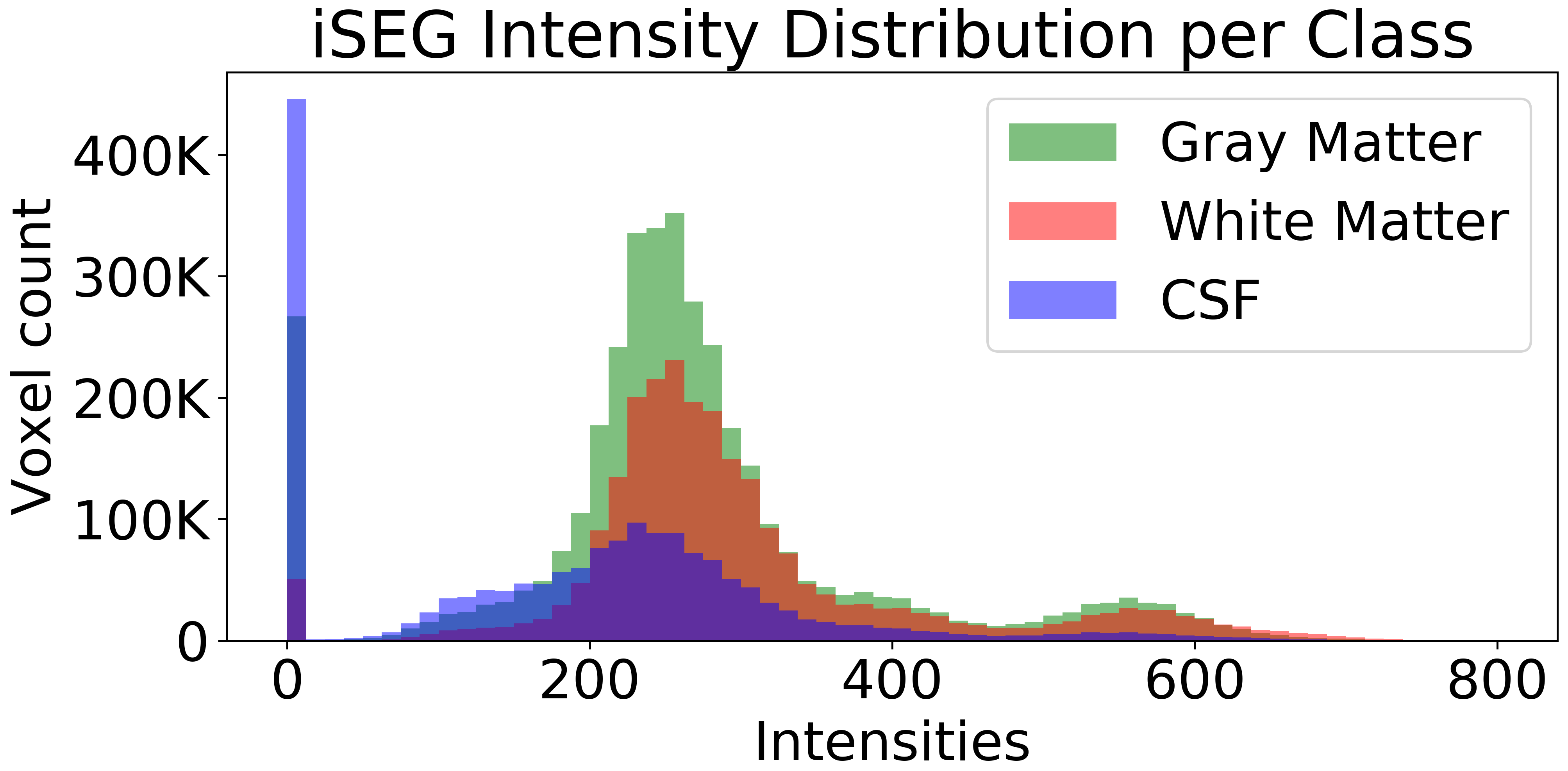}    
    \vspace*{-2mm}
    \caption{Intensity histograms of different brain tissue classes for adult brains in the MRBrainS dataset and 6-8 month infants in the iSEG dataset. We can see the important overlap in intensities, especially for iSEG, which is the cause of misleading classifiers. One can also notice the intensity range being largely different between both datasets.}
    \label{figure1}
\end{figure}

\subsection{Related Work}

\mypar{Image Normalization} 
A plethora of pre-processing techniques exists to normalize medical images prior to any image analysis. One common approach, known as standardization~\citep{Birenbaum2016, Kamnitsas2017, Casamitjana2016, Chen2018}, consists of normalizing each pixel intensity value in an input image by subtracting from it the image average intensity and dividing it by the its standard deviation. However, this simple strategy does not take into account the global statistics of the dataset. Other pre-processing approaches, such as histogram equalization~\citep{Onofrey2019} and bias field correction~\citep{Birenbaum2016, Baid2018, Feng2019}, are also commonly used to mitigate the problem of intensity inhomogeneity in images. For instance, \cite{Onofrey2019} evaluate the benefit of using different normalization techniques to multi-site prostate MRI before applying deep learning-based segmentation. Recently, a few studies have explored the potential of learning methods for dynamic data augmentation and normalization \citep{Drozdzal2018,Ciga2019,Hesse2020} as well as image denoising \citep{oguz2020}.  \cite{Drozdzal2018} use two consecutive fully-convolutional CNNs, a pre-processor network followed by a segmentation network trained with a Dice metric, to normalize an input image prior to segmentation. Despite showing a better performance compared to the segmentation of unnormalized images, this prior work has several limitations. First, since there is no realism constraint on the intermediate images produced by the pre-processing network, these images lack interpretability across multiple datasets. Moreover, because images are encoded for a specific network, they cannot be used with other segmentation models without retraining. Last, the feasibility of this approach was not demonstrated for multi-site settings where the multiple datasets are used jointly to learn the normalization. A domain adaptation strategy that adds a domain classifier at the end of each layer of a classification network to extract domain-invariant features is explored in~\citep{Ciga2019}. While it supports multiple domains, such as datasets acquired with different imaging protocols, this strategy is not tailored to a specific task like segmentation. Thus, it yields suboptimal results compared to a task-driven normalization approach. To this effect, we show in our experiments that learning a segmentation network jointly with the normalization network actually improves the contrast between different regions of interest (ROIs) in the normalized images. 

\mypar{Data Harmonization} The harmonization of data across multiple sites has also sparked interests in research, specially to increase the sample size of statistical studies~\citep{LOGUE2018244}. \cite{Shinohara2017} show that employing scanners from the same vendor and carefully harmonizing the protocols for the acquisition of multicenter 3D MRI brain data  still results in systematic image differences. This impacts the accuracy of volumetric analyses, notably introducing bias in measured white and gray matter volumes. Moreover, multiplying the number of sites can introduce nonlinear age-related differences in ROIs within the brain. \cite{pomponio2020harmonization} use data harmonization to remove site-related demographics effects in the cross-sectional LIFESPAN dataset. However, data harmonization is only done after the segmentation of ROIs, adding an extra step to the processing pipeline. Data harmonization approaches based on deep learning have also been explored in recent work. \cite{dewey2019deepharmony} proposed an FCN architecture based on 3D U-Net for contrast harmonization between two different protocols, demonstrating a more consistent volume quantification across these protocols. However, training this architecture requires paired images of the same subjects with different acquisition protocols, which is challenging to obtain in practice and does not scale to multi-site data. \cite{modanwal2020mri} use a cycle-consistent generative adversarial network (CycleGAN) to generate harmonized structural breast images between two different types of scanner. The proposed method leverages unpaired data with two generator-discriminator pairs to bypass common limitation of image translation algorithms that require paired data. This approach has also been explored for image denoising \citep{oguz2020}. While the results show visually-realistic harmonized images, it does not consider the specific image analysis task performed after pre-processing. Moreover, this harmonization approach is limited to only two domains, and extending it to additional ones requires substantial modifications. In contrast, our method accommodates an arbitrary number of data sites without added complexity. 

\section{Material and Methods}\label{sec:methods}

\subsection{The Proposed Architecture}

\begin{figure*}[t!]
    \begin{center}

        \includegraphics[width=.75\linewidth]{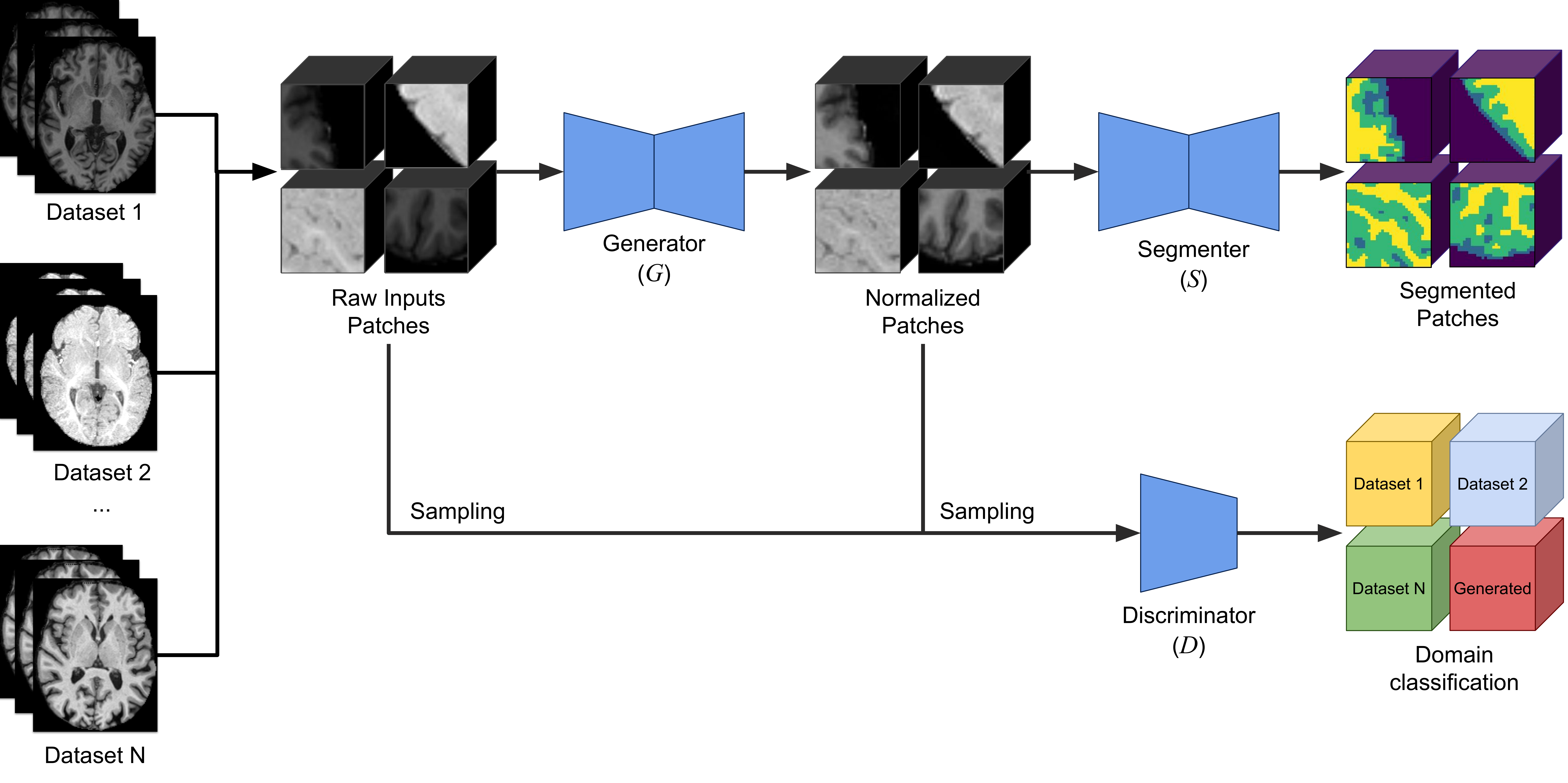}
        
        \vspace*{-2mm}
        \caption{Proposed architecture. A first FCN \emph{Generator} network ($G$) takes a non-normalized patch and generates a normalized patch. The normalized patch is input to a second FCN \emph{Segmenter} network ($S$) for proper segmentation. \emph{Discriminator} ($D$) network apply the constraint of realism on the normalized output. The algorithm learns the optimal normalizing function based on the observed differences between input datasets.}\label{fig:diagram}
    \end{center}
\end{figure*}

Our adversarial image normalization architecture, illustrated in Fig.~\ref{fig:diagram}, consists of three main components: 1) a fully-convolutional Generator ($G$) which acts as an image pre-processor to produce normalized images; 2) a Segmenter ($S$) based on the same fully-convolutional network that outputs a segmentation map from normalized images; 3) a Discriminator ($D$) which tries to classify the domain of normalized images. In the following, we describe each of these components in greater details.

\mypar{Generator} Although any other FCN architecture could be used, our Generator network $G$ is based on a modified 3D U-Net~\citep{Cicek} where the change resides primarily in the expanding path. The original 3D U-Net uses transposed 3D convolutions and concatenates feature maps to recover spatial resolution. Because of GPU memory constraints, we employ a simpler upscaling operator to upsample the resolution at each level of the U-Net decoder coupled with a feature concatenation. This enables the reduction of the total number of parameters in our model. We kept the same encoding path as in \citep{Cicek}, which consists of alternating convolutions and max-pooling layers. The network also uses shortcut connections from encoding to decoding path between layers of equal resolution to help recover the high-resolution features. Each convolution is followed by a batch normalization and a ReLU activation function. We also kept the same number of feature maps in each convolution. During training and testing, the model takes a 3D patch $\xx \in \real^{|\img|}$, where $\img$ is the set of patch voxels, and transforms it into a cross-domain normalized ima\-ge $\xnorm = G(\xx)$. 

\mypar{Segmenter} The segmentation network $S$ uses the same 3D U-Net architecture as our Generator. This network receives the normalized output of the image generator and performs voxel-wise classification using a final $1^3$ convolution before the softmax layer. The output of this network is the segmentation map of the input patch $S(\xnorm)$. While any other segmentation loss can be employed to train $S$, we chose the widely used Dice loss~\citep{milletari2016v, Carass2020} defined as 
\begin{equation}\label{eq:dice_loss}
    \lossSeg (\sss,\yy) \ = \ 1 \, - \, \frac{\epsilon \, + \, 2 \sum_{c} \, \omega_c \sum_{v \in \img} s_{v,c} \cdot y_{v,c}}{\epsilon \, + \, \sum_{c} \, \omega_c \sum_{v \in \img} (s_{v,c} +  y_{v,c})}, 
\end{equation}
where $s_{v,c} \in [0, 1]$ is the softmax output of $S$ for voxel $v$ and class $c$, $y_{v,c}$ is the corresponding ground-truth label (i.e., $y_{v,c}$ is 1 if the ground-truth class of voxel $v$ is $c$, else it is 0), and $\epsilon$ is a small constant to avoid zero-division. The weight $\omega_c$ controls the influence of class $c$ in the loss. It is typically tuned to alleviate the problem of class imbalance by giving a higher weight to smaller-region classes in images. Throughout all experiments, we empirically fixed these weights to $\omega_{\mr{\textsc{BG}}}$\,=\,0.22, $\omega_{\mr{\textsc{WM}}}$\,=\,0.28, $\omega_{\mr{\textsc{GM}}}$\,=\,0.20 and $\omega_{\mr{\textsc{CSF}}}$\,=\,0.30. 

\mypar{Discriminator} For the domain classifier $D$, we chose the DCGAN~\citep{Radford2016} discriminator's architecture with 5 layers and adapted its implementation to 3D volumes. The first 4 layers consist of a 3D convolutional operation, a LeakyReLU activation, and a dropout operation. A final linear layer follows to ensure classification. Note that we also tried the more recent ResNet model, which employs residual connections to improve gradient flow during training. However, we found that this model tends to overfit and leads to a worse performance than DCGAN's discriminator. Our final classification network receives as inputs both raw dataset patches and normalized patches from the Generator. Each patch has an image domain label $z \in \{1, \, \ldots, \,K\!\!+\!1\}$ which determines from which dataset the patch comes from (labels 1 to $K$) or if the patch is generated (label $K$+1). The role of this discriminator is to ensure that ima\-ges produced by $G$ are both realistic and domain-invariant. Although other classification losses could be considered, we used the negative log likelihood (cross entropy) loss, i.e., 
\begin{equation}\label{eq:discr_loss}
    \lossDis\big(D(\xx), \,z\big) \ = \ -\log \, D_z(\xx),
\end{equation} 
where $D_z(\xx)$ is the softmax probability for class $z$. We note that, since $\sum_{z=1}^{K+1}D_z(\xx)=1$, the loss for the ($K$+1)-th class corresponding to generated examples can be written in terms of domain classes as
\begin{equation}
    \lossDis\big(D(\xx), \,K\!\!+\!1\big) \ = \ - \log \Big(1 - \sum_{z=1}^K D_z(\xx)\Big)\nonumber. 
\end{equation}

{\centering
\SetAlFnt{\small}

\begin{algorithm2e}[t!]

\SetNoFillComment

\KwIn{Training set $\data = \{(\xx_i, \yy_i, z_i\}_{i=1}^{|\data|}$}
\KwIn{Batch size $m$, number epochs, iterations and steps ($n_{epochs}$, $n_{iter}$, $n_{steps}$), and learning rates $\eta_G$, $\eta_S$, $\eta_D$;}
\KwOut{Network parameters $\params_G$, $\params_S$, $\params_D$;}
\caption{Training of our adversarial image normalization method, with the interacting updates of the Discriminator, the Segmenter and the Generator. 
}\label{algo}

    \BlankLine
    Randomly initialize network parameters $\params_G$, $\params_S$, $\params_D$\;
    \BlankLine
    \For{${epoch} = 1,\ldots,n_{epochs}$}{ %\Do
    \For{${iteration} = 1,\ldots,n_{iter}$}{ %\Do
        \For{$step = 1,\ldots,n_{steps}$}{ %\Do
            Sample batch of $m$ examples from all domains $\{ (\xx_i, z_i)\}_{i=1}^m$\;

            Update the \textcolor{myred}{Discriminator $D$}:  
            \begin{empheq}[box=\fcolorbox{myred}{white}]{align*}
            %\begin{align*} 
            \params_D & \ \gets \ \params_D \, - \, \frac{\eta_D}{m} \sum_{i=1}^{m}\Big( \nabla_D\,\lossDis\big(D(\xx_i), z_i\big)\\[-3pt]
            & \qquad\qquad\qquad \, + \, \nabla_D\,\lossDis\big(D(G(\xx_i)), K\!\!+\!1\big)\Big);
            %\end{align*}
            \end{empheq}
        }
        
        Sample batch $m$ examples from all domains $\{ (\xx_i, \yy_i)\}_{i=1}^m$\;     
        Update the \textcolor{mygreen}{Segmenter $S$} and \textcolor{blue}{Generator $G$}:
        %\begin{align*}
        \begin{empheq}[box=\fcolorbox{mygreen}{white}]{align*}
        & \params_S \ \gets \ \params_S \, - \, \frac{\eta_S}{m} \sum_{i=1}^{m} \nabla_S\,\lossSeg\big(S(G(\xx_i)), \yy_i\big);
        \end{empheq}
        \begin{empheq}[box=\fcolorbox{blue}{white}]{align*}
        & \params_G \ \gets \ \params_G \, - \, \frac{\eta_G}{m} \sum_{i=1}^{m}\Big(\nabla_G\, \lossSeg\big(S(G(\xx_i)), \yy_i\big) \\[-3pt]
        & \qquad\qquad\qquad  \,  - \, \lambda\nabla_G\,\lossDis\big(D(G(\xx_i)), K\!\!+\!1\big)\Big); % ;)}
        %\end{align*}        
        \end{empheq}
    }
    Adjust learning rates $\eta_G$, $\eta_S$, $\eta_D$\;
    }
\Return{$\params_G$, $\params_S$, $\params_D$} \;
\end{algorithm2e}
}

\subsubsection{Adversarial Training}

The three networks of our model are trained together in an adversarial manner by optimizing the following loss function: 

\begin{equation}
\begin{aligned}
    \min_{G,S} \ \max_{D} \ \loss(G, S, D) & \ = \ \expect_{\xx,\yy } \, \big[\lossSeg \big(S(G(\xx)), \yy\big)\big]  \\ 
    & - \lambda\, \expect_{\xx,z} \Big[ \lossDis \big ( D(\xx), z \big) \, + \, \lossDis \big ( D \big( G(\xx) \big ), K\!\!+\!1 \big) \big ]
    \label{eq:total_loss}
\end{aligned}    
\end{equation}

where $\lossSeg$ and $\lossDis$ are the segmentation and discriminator losses, respectively defined in Eq.~(\ref{eq:dice_loss}) and Eq.~(\ref{eq:discr_loss}). Hyper-parameter $\lambda$ controls the trade-off between having a good segmentation accuracy (first loss term) and having normalized images which are domain-invariant (last loss term). By using $\lambda$\,=\,0, the model becomes similar to \citep{Drozdzal2018}, where the generator is not constrained to produce realistic images. In contrast, for a large $\lambda$, our model becomes similar to an adversarial domain classifier~\citep{Ciga2019}, where generated images are normalized across different domains but not optimal for segmentation, with added realism constraints.

The training procedure of our method is detailed in Algorithm \ref{algo}. As in standard adversarial learning approaches, we train our model by updating the generator and discriminator in two separate steps. 
The discriminator is updated $n_{steps}$\,=\,3 times to maintain near optimal solution of domain classification while $G$ is updated less frequently. The segmentation network is then updated at the same frequency as $G$. We adopt a mini-batch stochastic gradient descent (SGD) technique and use Adam ~\citep{Kingma2014} optimizer to update parameters at each step, where the gradient is estimated using a batch of $m$ training examples, and the update step size is controlled by learning rate $\eta_G$, $\eta_S$, $\eta_D$.

Although it is possible to enforce the realism and domain invariance of normalized images via two separate discriminators, employing a single discriminator provides several important advantages. First, it avoids the problem of instability which results from training discriminators with competing losses. Instead of treating image domain and realism as unrelated properties, our model predicts them jointly within a single network. Moreover, our single discriminator model has fewer hyper-parameters to tune and is less expensive in terms of computation and memory. In addition to its higher simplicity, our adversarial model without the segmentation loss can also be shown, under mild assumptions, to lead to the desirable solution where normalized images are generated from the mean distribution of real images. This is done in the following theorem.

\begin{thm}\label{th1}
Let $p_r(\xx \, | \, z)$ and $p_g(\xx \, | \, z)$ be the probabilities that $\xx$ is a real or a generated image, respectively, from source dataset $z$. The minimax optimization problem of Eq. (\ref{eq:total_loss}) without the segmentation term corresponds to minimizing the divergence between $p_g(\xx \, | \, z)$ for each $z$ and the mean distribution of real images $\overline{p}_r(\xx) = \frac{1}{k}\sum_{z=1}^K\,p_r(\xx \, | \, z)$.
\end{thm}

\begin{pf} See Appendix \ref{sec:proof}.
\end{pf}

\subsection{Data and performance metrics}\label{sec:data}

To evaluate the performance of our method, we selected three databases with important differences in their intensity profile and subject demographics. 
%\begin{itemize}\setlength\itemsep{.1em}
\mypar{iSEG} The first dataset, iSEG~\citep{Wang2019}, comprises 10 T1 and T2 MRI data of 6-8 month old infants acquired with a 3 Tesla scanner. The ground truth is the segmentation mask of the three main brain tissues, white matter (WM), gray matter (GM) and cerebrospinal fluids (CSF), which are critical for detecting abnormalities in brain development. Images are sampled into an isotropic 1.0\,mm$^3$ resolution. This dataset is particularly challenging for segmentation since subjects are in an isointense phase where the white matter and gray matter voxel intensities greatly overlap, thus leading to a lower tissue contrast. Its images are also noisier because of the shorter scanning time used to avoid motion artifacts. 

\mypar{MRBrainS} The MRBrainS13~\citep{Mendrik2015} dataset contains 5 healthy adult subjects with T1 and T2 FLAIR modalities. Images were acquired from a 3 Tesla scanner following a voxel size of 0.958\,mm \ttm{}  0.958\,mm \ttm{} 3.0\,mm. This dataset has the same ground-truth classes as iSEG.

\mypar{ABIDE} The Autism Brain Imaging Data Exchange (ABIDE I)~\citep{DiMartino2014} was also used to further validate our method on an independent multi-site dataset. It comprises 1,112 images of normal and autism spectrum disorder (ASD) subjects. Images were acquired across 17 international sites, thus providing a high variance in intensity distribution. The anatomical scan parameters for each site are available on the ABIDE website\footnote{\url{http://fcon_1000.projects.nitrc.org/indi/abide/}}. A different source of variability in this dataset comes from the broad age span of its subjects, ranging from 7 to 64 years. The detailed demographics of subjects are shown in Table \ref{tab:abide}. Note that the data of 9 subjects were excluded from our study due to poor image image quality (e.g., important motion artifacts). Since this dataset lacks the manual segmentation masks, we instead considered the segmentation maps produced by the FreeSurfer \emph{recon-all}\footnote{\url{https://surfer.nmr.mgh.harvard.edu/fswiki/recon-all}} pipeline as ground-truth.
%\end{itemize}

\begin{table}[t!]
\centering
\caption{Demographics of subjects in the ABIDE dataset~\citep{DiMartino2014}.}
\label{tab:abide}
\begin{footnotesize}
\renewcommand{\arraystretch}{1.05}
\begin{tabular}{lcccc}
\toprule
\multirow{2}{*}{\B Group} & 
\multirow{2}{*}{\B n} & \multirow{2}{*}{\B Male} & \multirow{2}{*}{\B Female} & \B Age \\
& & & & \B (mean $\pm$ stdev) \\
\midrule
Control & 539 & 474 & 65 & 17.01 $\pm$ 8.36 \\
ASD & 573 & 474 & 99 & 17.08 $\pm$ 7.72\\
%\midrule
% Total & 1,112 & 948 & 164 & \\
\bottomrule
\end{tabular}
\end{footnotesize}
\end{table}

For iSEG, 8 subjects were randomly selected for training, while another was kept for validation and the remaining one was kept for testing. For MRBrainS, 3 images were randomly selected for training, 1 image was kept for validation and the last one for test purposes. For ABIDE, which has more images available for learning, all 1,103 usable images were randomly split in three sets: 60\% for training, 20\% for validation, and 20\% for testing. 

We evaluate the segmentation performance of tested methods using the mean Dice Similarity Coefficient (DSC), which measures the degree of overlap between the predicted segmentation map $\vec{S}$ and the ground-truth $\vec{G}$: 
\begin{equation}
     \mr{DSC}(\vec{S},\vec{G}) \ = \ \frac{2\,|\vec{S} \cap \vec{G}|}{\,|\vec{S} \cup \vec{G}|}.
\end{equation}
We also consider the Mean Hausdorff Distance (MHD) to measure the segmentation quality in terms of its boundary: %: \\
\begin{equation}
\mr{MHD}(\vec{S}, \vec{G}) \ = \ \frac{1}{2}\big(\mr{dist}(\vec{S}, \vec{G}) \, + \, \mr{dist}(\vec{G}, \vec{S})\big).
\end{equation}
Here, $\mr{dist}(\cdot,\cdot)$ is the maximum Euclidean distance between a point in the predicted segmentation map and its nearest point in the ground-truth (or vice-versa). 

\begin{figure}[t!]
\centering
    \begin{footnotesize}
    \mbox{
    \shortstack{\includegraphics[width=.245\linewidth]{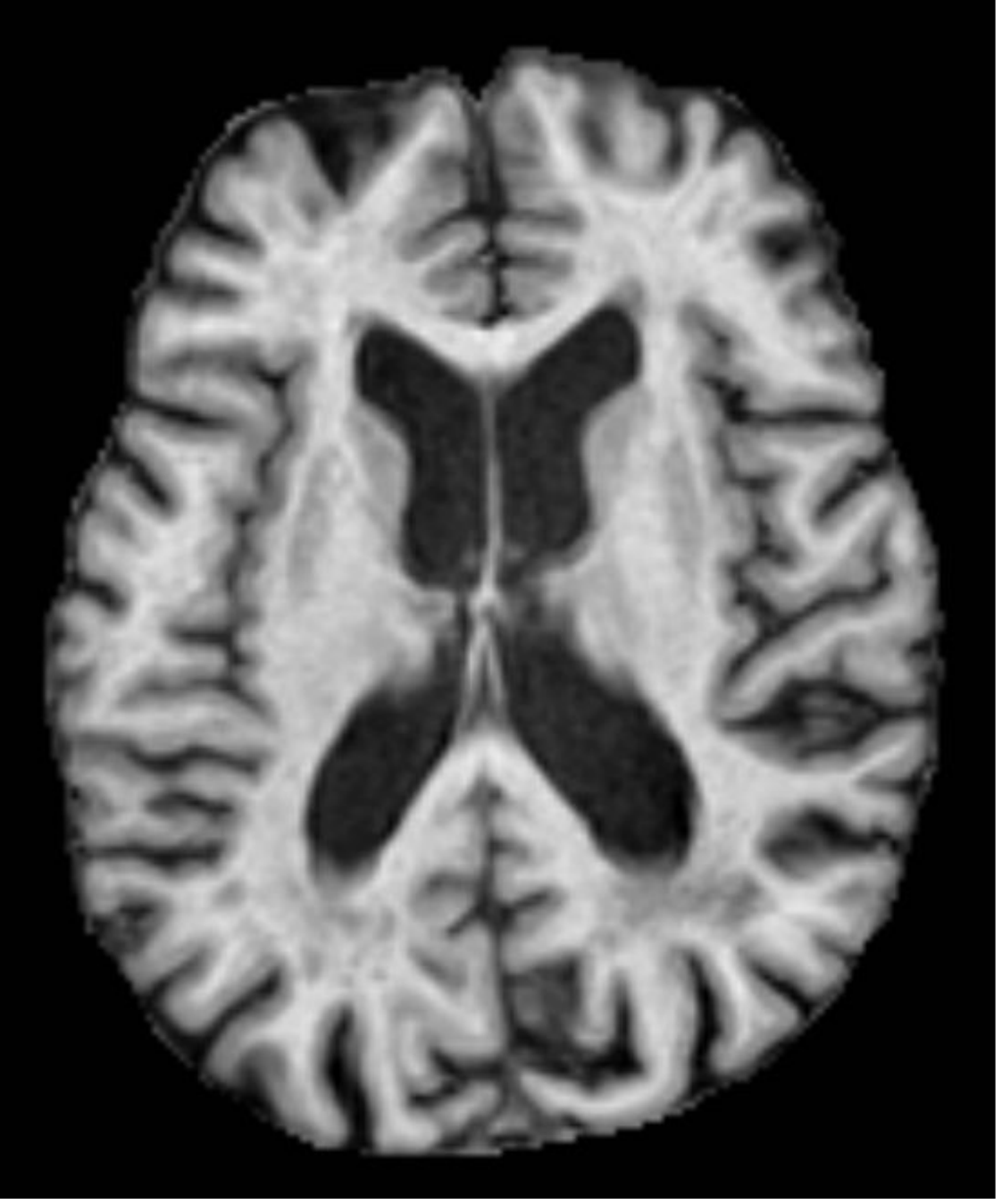} \\ Input}
    \shortstack{\includegraphics[width=.245\linewidth]{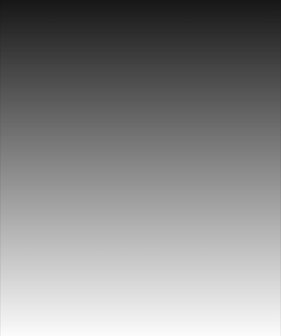} \\ Bias field}
    \shortstack{\includegraphics[width=.245\linewidth]{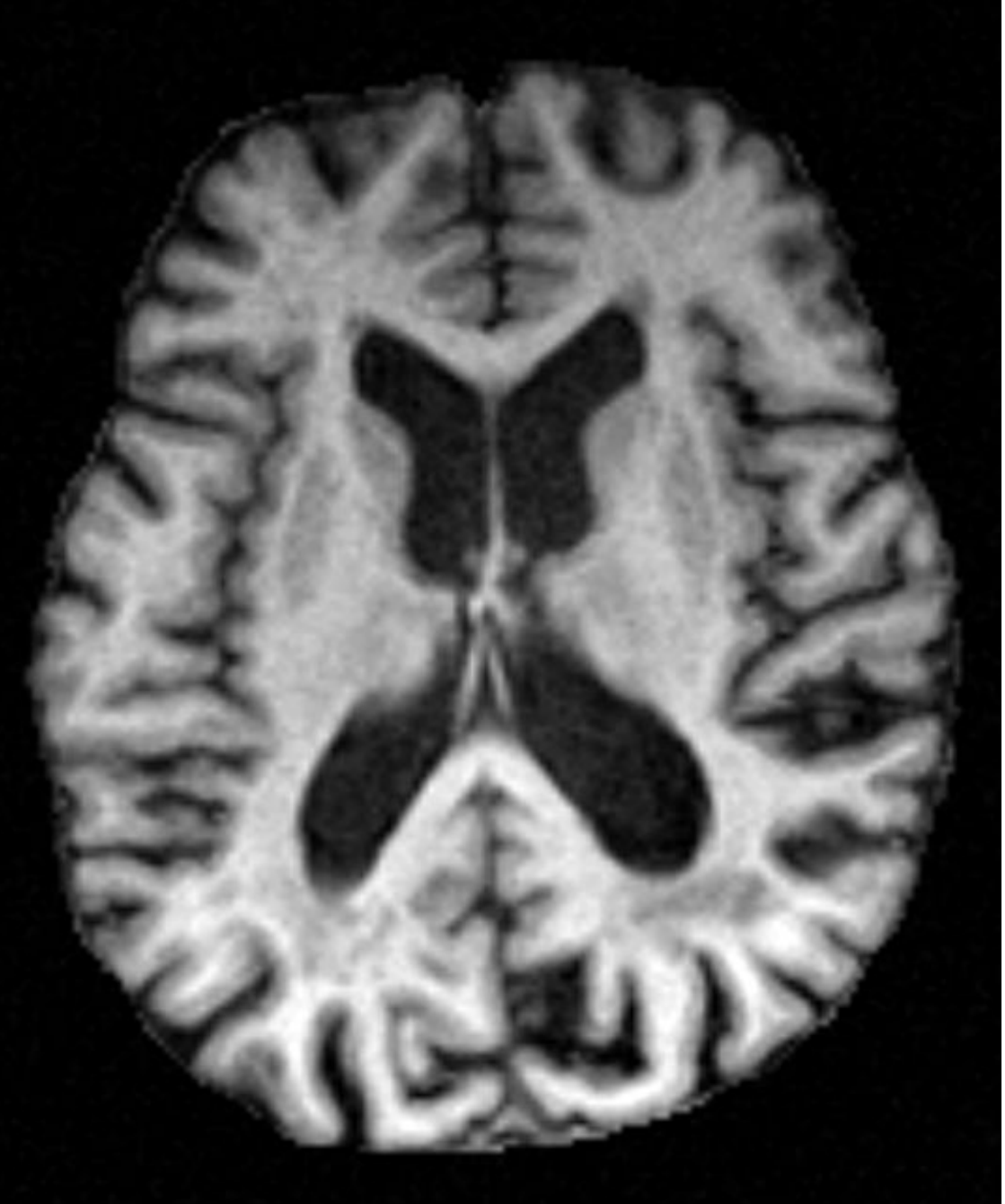} \\ Degraded image}
    \shortstack{\includegraphics[width=.245\linewidth]{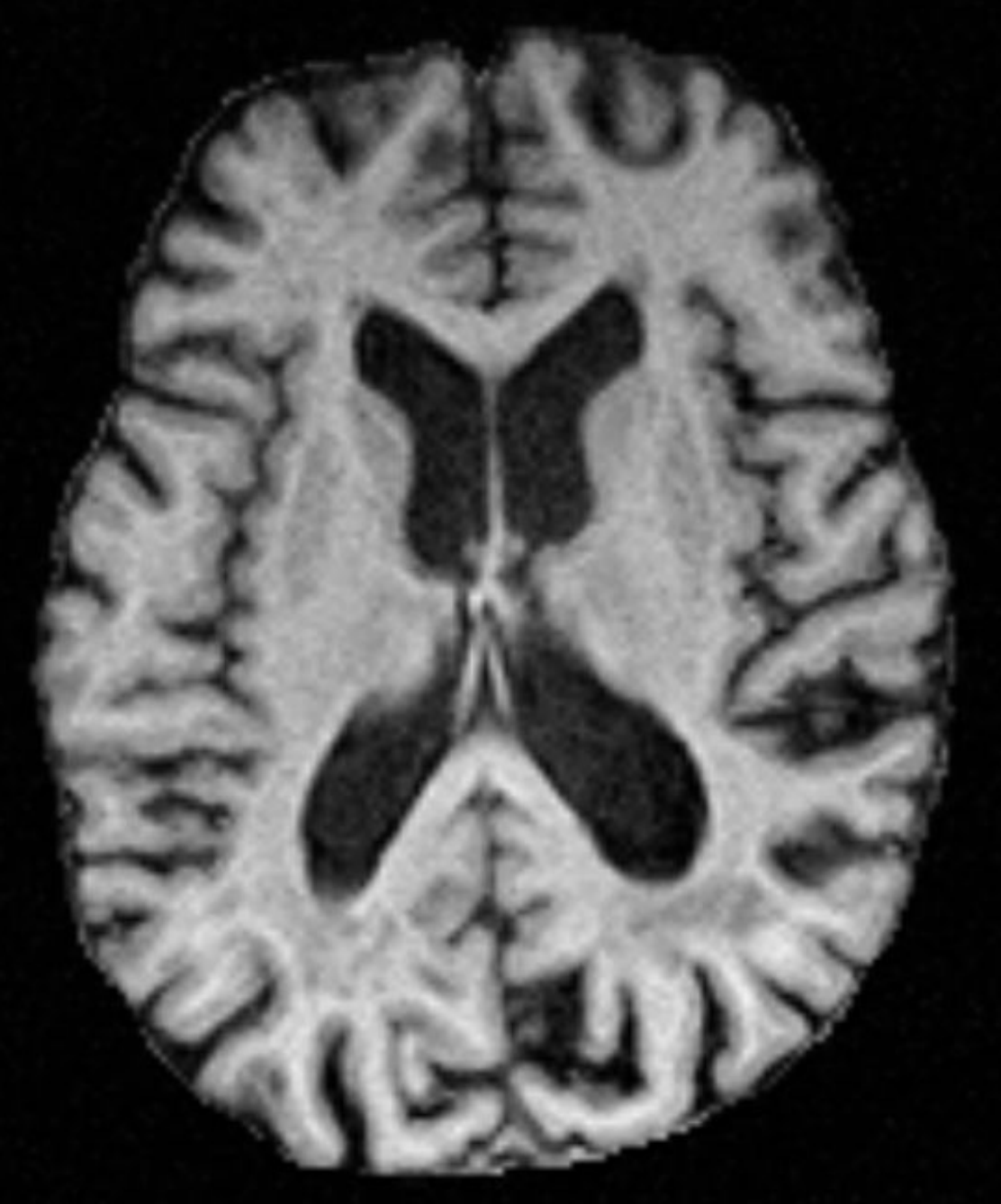} \\ Normalized}
    }
    \end{footnotesize} 
    
    \vspace*{-2mm}
    \caption{Normalization of an image degraded with bias field of strength $\alpha$\,=\,0.5. \textbf{Left}: Input test image. \textbf{Middle-left}: Applied bias field. \textbf{Middle-right}: Resulting degraded test input. \textbf{Right}: Normalized image produced by the generator. One can notice the more uniform distribution of intensities in the normalized image.}
    \label{fig:example_degraded}
\end{figure}

\subsection{Pre-processing}\label{sec:pre-processing}

For all three datasets, images were cropped to brain dimensions and then padded with zero-intensity voxels to have a fixed size corresponding to the largest brain. Skull stripping was performed using the segmentation map of MRBrainS. Images were resampled to be 1.0\,mm$^3$ isotropic in order to match the iSEG resolution. Due to limited GPU memory, volumes were processed (normalization and segmentation) in separate, overlapping 3D patches of size $32^3$ voxels. To train the model, 40,000 patches are randomly sampled among training images of each dataset, all centered on a voxel of a foreground class. Likewise, 12,000 patches are sampled from validation images in each dataset to evaluate the model at each epoch. The same subjects and patches were selected across all experiments through a common random seed. For testing, we process full volumes in $32^3$ sub-patches extracted with a stride of $8^3$ voxels. Whole-image segmentation is obtained by averaging the class probabilities of each voxel across patches containing this voxel. 

To test the robustness of our model to intensity inhomogeneity, we apply a data augmentation strategy where input patches have a 50\% probability of being transformed with a multiplicative bias field. Let $I(x,y,z)$ be the intensity of the input image at voxel $(x,y,z)$, the  transformed image can be defined as in \citep{song2017review}: 
\begin{equation}
I'(x,y,z) \ = \ I(x,y,z)\,B(x,y,z) \ + \ \eta(x,y,z)
\end{equation}
where $B(x,y,z)$ is the bias field at $(x,y,z)$ and $\eta(x,y,z)$ is additive noise at the same voxel. In our experiments, we implemented a simple noiseless model where intensities are scaled linearly along the y-axis:
\begin{equation}
B(x,y,z) \ = \ \frac{y}{H}\,\alpha \ + \ (1\!-\!\alpha), \ \ \alpha \in [0, 1].
\end{equation}
In the above expression, $\alpha$ is the coefficient defining the slope of the bias field and $H$ is the height of the image. While $\alpha$\,=\,0 corresponds to having no bias field, $\alpha$\,=\,1 scales intensities from $-$100\% when y\,=\,0 to +100\% when y\,=\,H.  An example of a transformed test image is shown in Fig. \ref{fig:example_degraded}. Note that the bias field slope for each data augmentation is randomly selected based on a uniform distribution. 

\subsection{Implementation Details}

Training was performed on an NVIDIA Tesla V100 32 GB GPU, with a total of 120 epochs for the dual dataset configurations (iSEG and MRBrainS) and 70 epochs when training with three datasets (iSEG, MRBrainS and ABIDE). 
All experiments were run to find the best ratio between the segmentation loss and discriminator loss, controlled by the hyper-parameter $\lambda$. For the main results, we selected a value of $\lambda$\,=\,1.5 which gave a good segmentation accuracy on the validation examples, while giving plausible generated images. We used a weight decay of 0.001 on all experiments. For the generator and segmentation network, we initialized the learning rate to $0.001$ and updated it with a multi-step strategy, decreasing it by a factor of 10 at epochs 50 and 75. Since the discriminator takes longer to train, a different strategy was adopted. Starting with a learning rate of $0.0001$, we instead used a reduce-on-plateau strategy which applies a decay of 0.1 when the validation loss does not decrease for 7 consecutive epochs. We trained each model with an Adam optimizer. The model was implemented using the PyTorch\footnote{\url{http://pytorch.org}} deep learning framework.

\section{Experiments and Results}\label{sec:experiments}

A series of experiments is conducted to assess the benefits of the proposed adversarial normalization method. We start by showing the poor generalization performance of a baseline segmentation model without normalization, when trained and tested on different datasets. We then evaluate our method in a dual-site setting with the iSEG and MRBrainS datasets, and compare its segmentation and normalization performance against that of two competing approaches. Afterwards, we demonstrate our method in a multi-site setting by adding the large-scale ABIDE dataset and test it on multi-modal data. Last, we show its ability to correct intensity inhomogeneity and evaluate the impact of hyper-parameter $\lambda$ on performance.

\begin{table}[t!]
  \centering
  \caption{Dice scores when using different training and testing sets, without any normalization. The non-adaptation of the acquisition domains seriously impacts the segmentation accuracy.}
    \label{tab:baseline}
  \begin{footnotesize}
  \renewcommand{\arraystretch}{1.05}
  \begin{tabular}{cccccc}
    \toprule
    &  & \multicolumn{4}{c}{\B Dice} \\
     \cmidrule(l{6pt}r{6pt}){3-6}
    \B Training & \B Testing & \B CSF & \B GM & \B WM & \B Mean \\
    \midrule
    iSEG & iSEG & 0.920 & 0.857 & 0.828 & 0.868 \\
    MRBrainS & MRBrainS & 0.861 & 0.789 & 0.839 & 0.830 \\
    iSEG & MRBrainS & 0.401 & 0.354 & 0.519 & 0.425\\
    MRBrainS & iSEG & 0.293 & 0.082 & 0.563 & 0.313\\
    \bottomrule
 
  \end{tabular}
  \end{footnotesize} 
\end{table}

\subsection{Cross-dataset Baseline Performance}

We first establish a baseline measuring the cross-dataset performance of our segmentation network without any normalization. For this experiment, we use the same 3D U-Net architecture as in our adversarial method and train it on unnormalized T1 images from different datasets using a Dice loss. Four different scenarios are tested: 1) training and testing on iSEG data only; 2) training and testing on MRBrainS data only; 3) training on iSEG and testing on MRBrainS; 4) training on MRBrainS and testing on iSEG. The last two scenarios evaluate the segmentation model's ability to generalize across datasets with different characteristics, including subject demographics and acquisition protocol. 

Results of this experiment are shown in Table \ref{tab:baseline}. As can be seen, the performance decreases considerably when testing on a different dataset from the one used for training. Thus, the mean Dice score of the model trained on iSEG drops by 51.0\% when tested on MRBrainS. Likewise, we observe a 62.3\% drop in mean Dice when testing on iSEG the model trained on MRBrainS. These results demonstrate the high sensitivity of deep learning segmentation models to the training data, thus validating the need for a data-driven normalization method.

\begin{table*}[t!]
%\begin{adjustwidth}{-7em}{-5em}
  \centering
  \caption{Comparative Dice scores of different model architectures and data. The proposed method yields an important performance improvement over training and testing on single-domain or on standardized inputs.}
\label{tab:comparison}
\begin{footnotesize}
  %\resizebox{\columnwidth}{!}{
  \renewcommand{\arraystretch}{1.05}
  \begin{tabular}{cccccccccc}
    \toprule
   \multirow[b]{2}{*}{\B Setting} & \multirow[b]{2}{*}{\B Method} & \multicolumn{2}{c}{\B CSF} & \multicolumn{2}{c}{\B GM} & \multicolumn{2}{c}{\B WM} & \multicolumn{2}{c}{\B Mean} \\
     \cmidrule(l{6pt}r{6pt}){3-4} \cmidrule(l{6pt}r{6pt}){5-6} \cmidrule(l{6pt}r{6pt}){7-8} \cmidrule(l{6pt}r{6pt}){9-10}
     &  & \B DSC & \B MHD & \B DSC & \B MHD & \B DSC & \B MHD & \B DSC & \B MHD \\
    \midrule
    \multirow{3}{*}{\shortstack{\textbf{Dual-site} \\[3pt] (iSEG\,+\,MRBrainS)}} 
    & Standardization & 0.897 & 0.792 & 0.836 & 0.498 & 0.790 & 0.734 & 0.841 & 0.675 \\     
     & Pre-processor & \B 0.919 & \B 0.227 & \B 0.860 & 0.517 & 0.831 & 0.702 & \B 0.870 & 0.482 \\
   &  Ours & 0.912 & 0.245 &  0.853 & \B 0.492 &  \B 0.836 & \B 0.595 & 0.867 & \B 0.444 \\
    % & \B Adversarial (ours) & \B 0.910 & \B 22.44 & \B 0.889 & \B 31.09 & \B 0.862 & \B 34.55 \\
     \midrule
    \multirow{3}{*}{\shortstack{\textbf{Multi-site} \\[3pt] (iSEG\,+\,MRBrainS\\ \,+\,ABIDE)}}  & Standardization & 0.860 & 0.264 & 0.881 & 0.684 & 0.856 & 0.812 & 0.866 & 0.587 \\
     & Pre-processor & \B 0.922 & \B 0.251 & \B 0.895 & \B 0.392 & 0.870 & \B 0.530 & \B 0.896 & \B 0.390 \\
     &  Ours &  0.913 & 0.293 & 0.887 & 0.422 & \B 0.870 & 0.598 & 0.890 & 0.438 \\
    \bottomrule
  \end{tabular}
  %}
  \end{footnotesize}
  %\end{adjustwidth}  
\end{table*}

\begin{figure}[t!]
\centering
    \begin{footnotesize}
    \setlength{\tabcolsep}{-0.5pt}
    \begin{tabular}{ccc}
    \figNorm{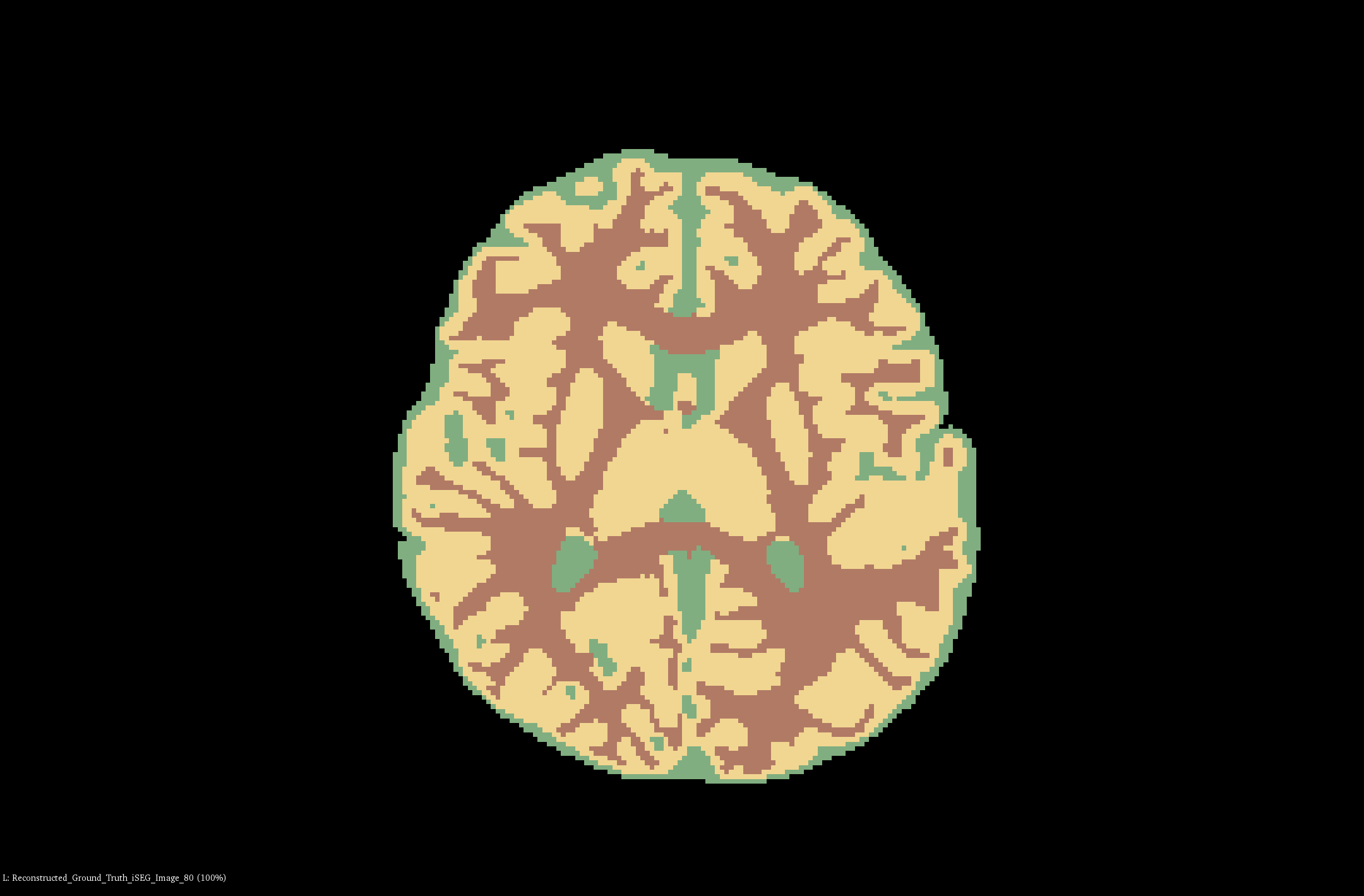} & \figNorm{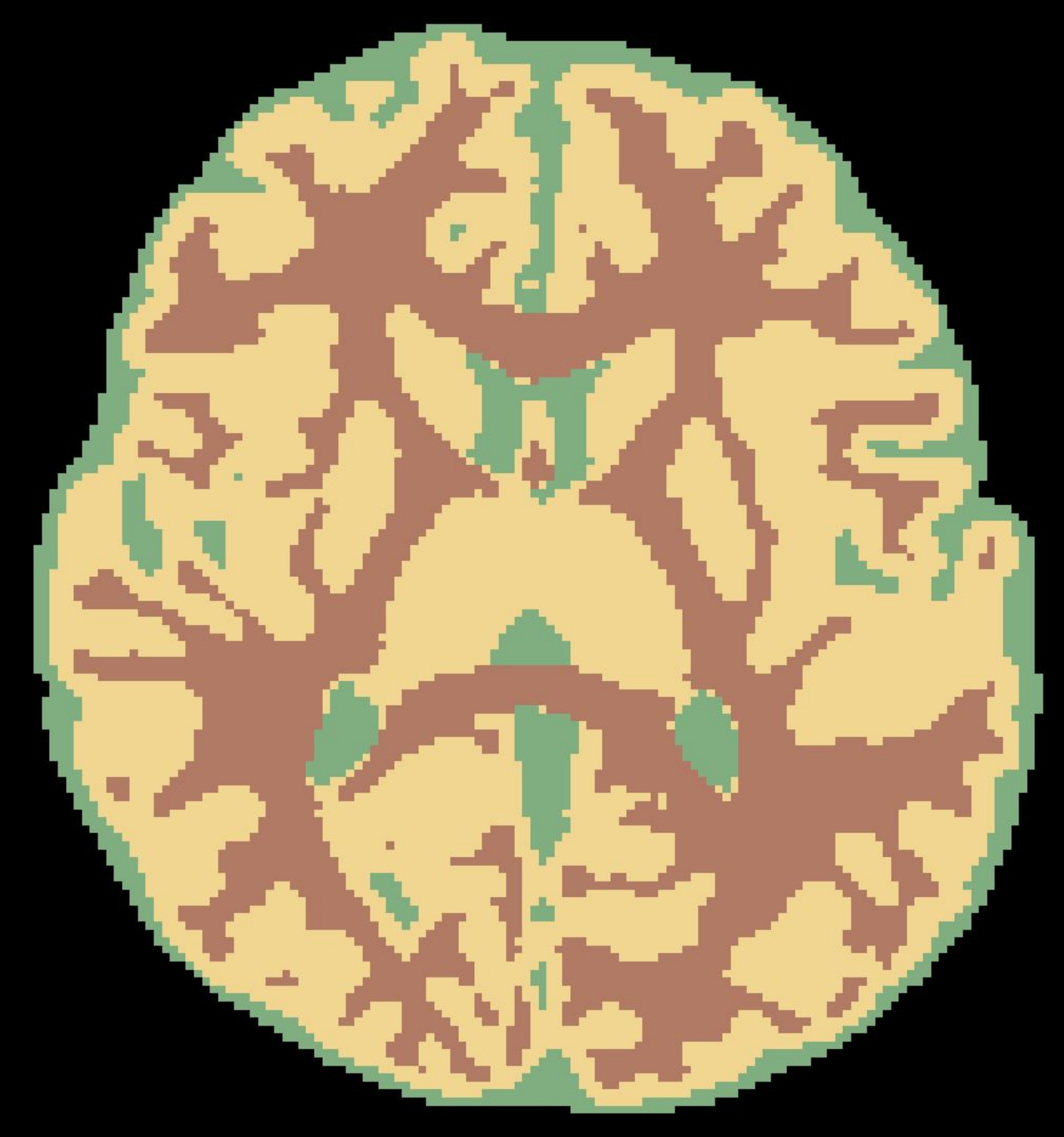} & \figNorm{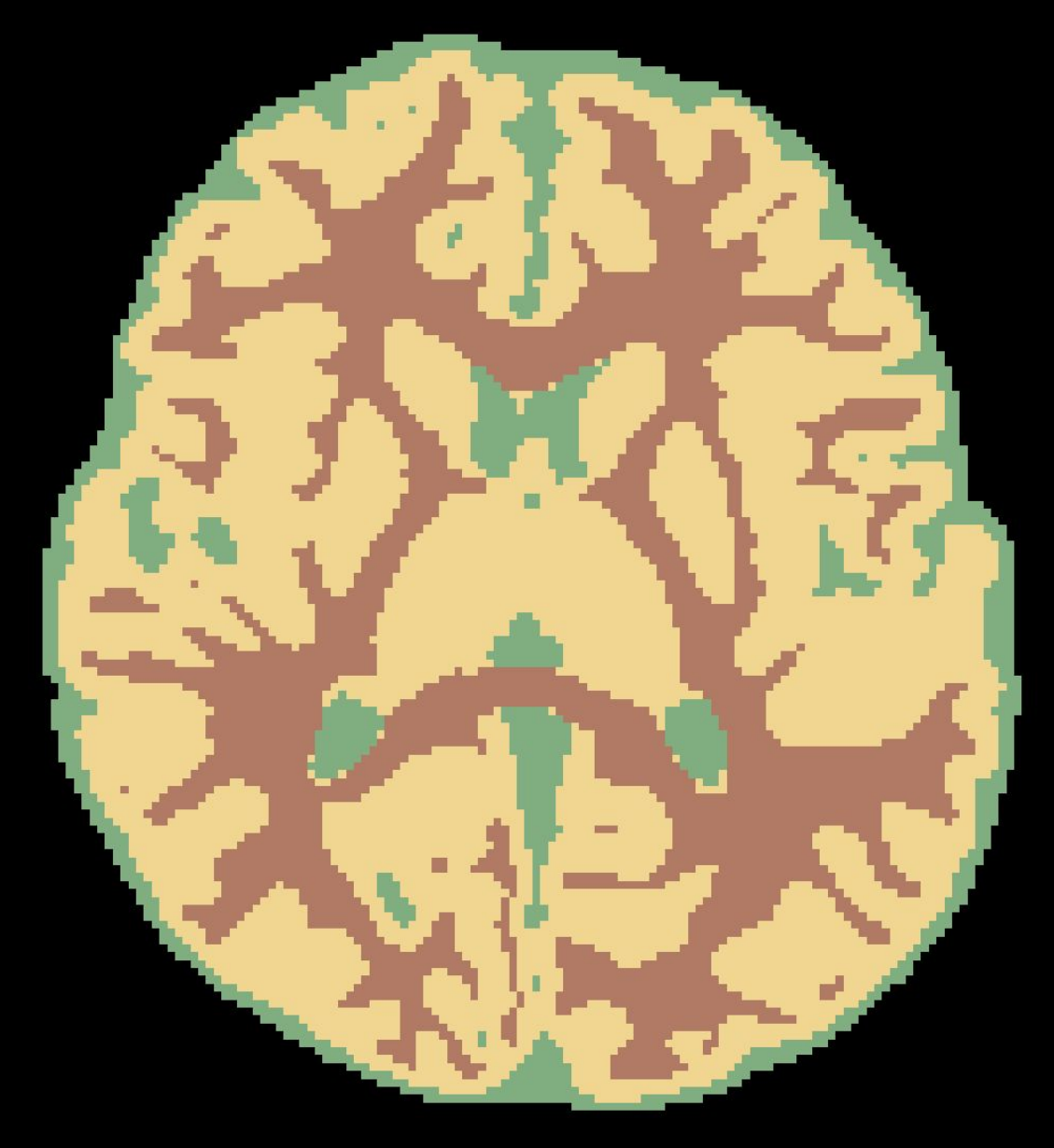} \\ 
    Ground-truth &
    Pre-processor%~\citep{Drozdzal2018} 
    &
    Ours \\
    \end{tabular}
    \begin{tabular}{ccc}
    \setlength{\tabcolsep}{-0.5pt}
    \figNorm{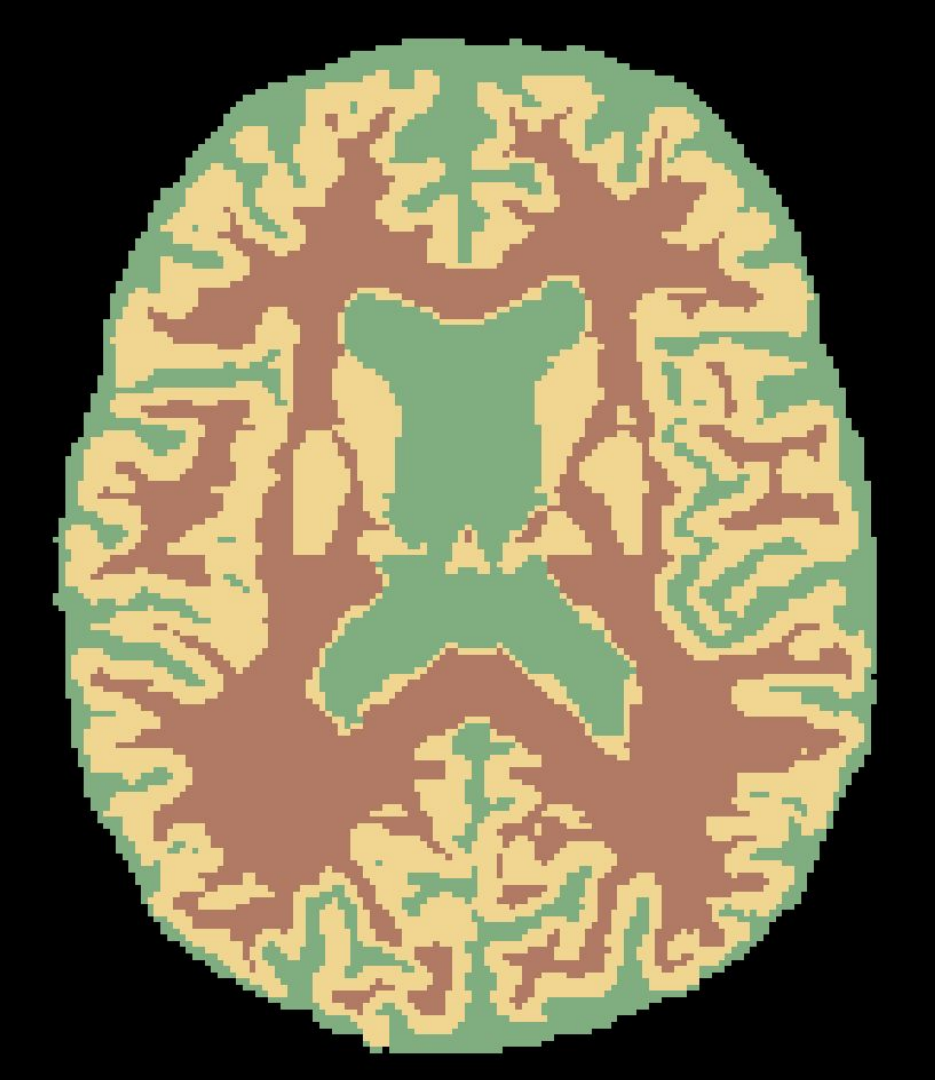} &  \figNorm{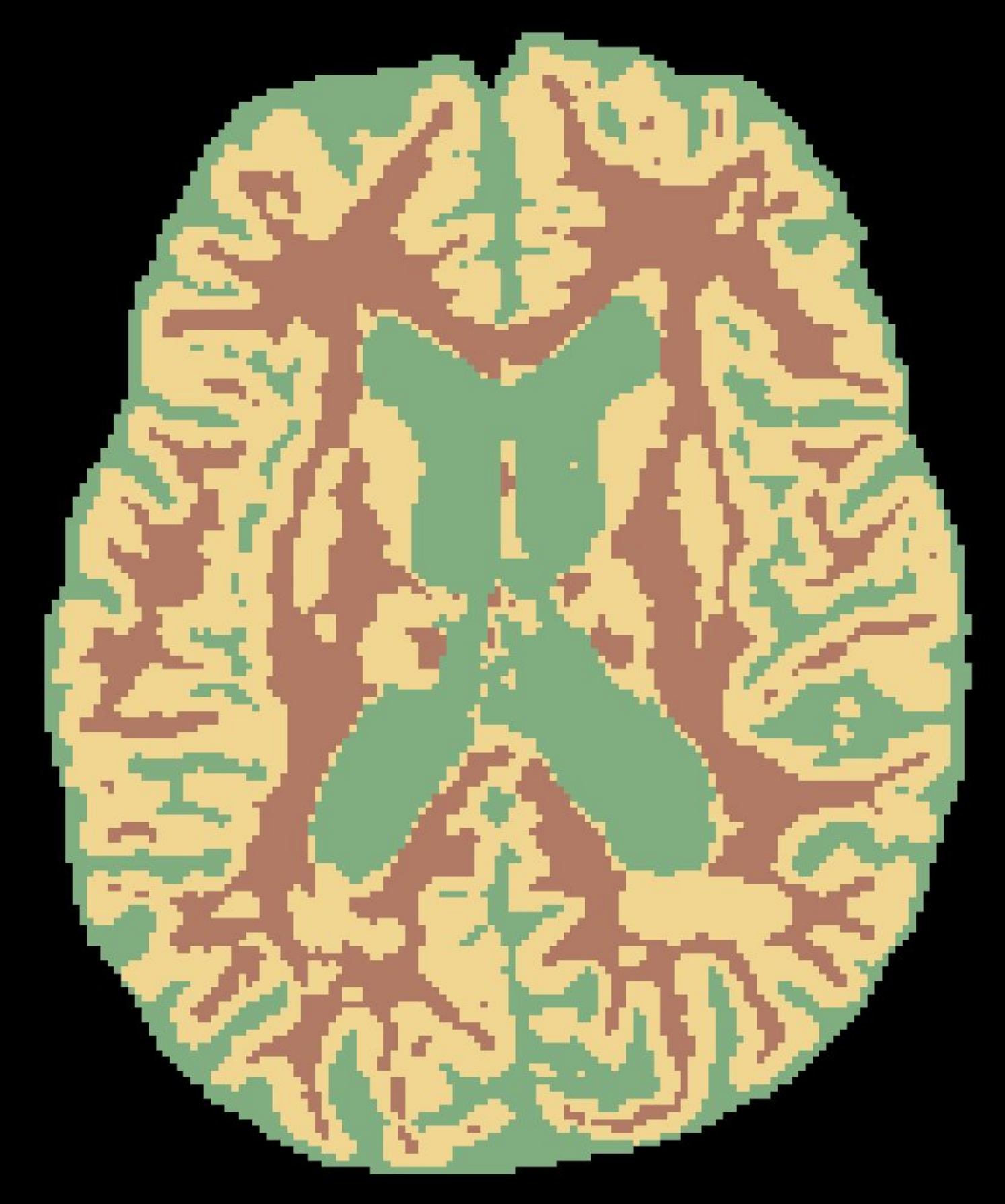} &
    \figNorm{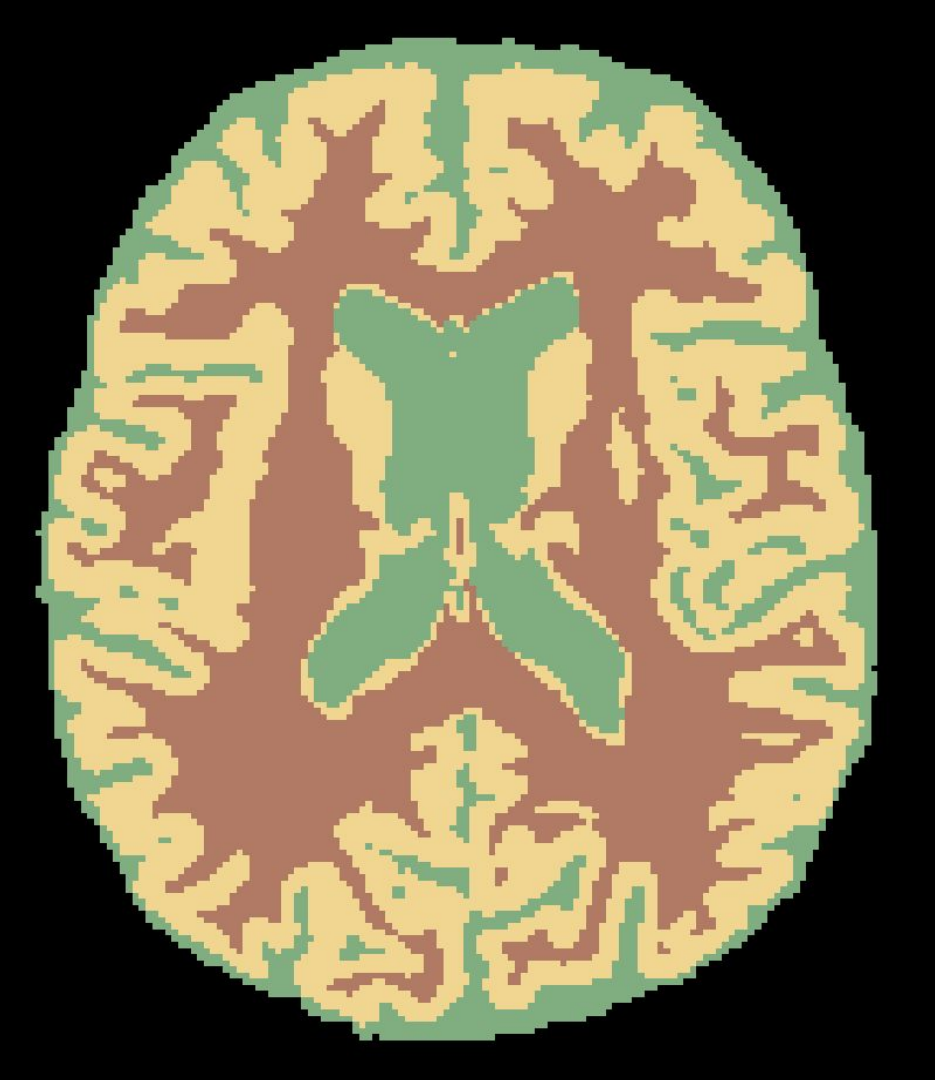} \\
    Ground-truth & 
    Pre-processor%~\citep{Drozdzal2018} 
    & 
    Ours
    \end{tabular}
    \end{footnotesize} 
    
    \vspace*{-2mm}
    \caption{Visualization of predicted segmentation masks for two test images. \textbf{Top row}: iSEG image. \textbf{Bottom row}: MRBrainS image. 
    }
    \label{fig:visualization}
\end{figure}

\subsection{Dual-site Evaluation}

Next, we compare our adversarial normalization method against a com\-mon\-ly-used standardization technique~\citep{Birenbaum2016} and the learned normalization approach of \cite{Drozdzal2018}, on images from two different datasets: iSEG and MRBrainS. As mentioned earlier, these datasets have distinct characteristics, iSEG containing T1 MRIs of infants in the isointense stage and MRBrainS the T1 MRIs of adult subjects. Therefore, standard per-image normalization may not be effective. The standardization technique tested in this experiment normalizes the intensity of each voxel in a given volume by subtracting from it the volume mean and dividing it by the standard deviation. The learned normalization approach, which we call \emph{Pre-processor} in the results, contains the same pipeline as our method (i.e., generator and segmentation networks), but without its discriminator. This baseline is used to asses the contribution of the adversarial learning in obtaining realistic normalized images. As this segmentation-optimized approach does not impose any constraint on realism, it should be an ``upper-bound'' on the segmentation accuracy achievable by our method.  

\subsubsection{Segmentation Performance}

The Dice scores obtained by the Standardization technique, the Pre-processor without realism constraints and our adversarial normalization me\-thod are reported in Table \ref{tab:comparison} for the iSEG and MRBrainS datasets
(dual-site setting). As expected, both our method and the learned pre-processor yield a large gain in performance over the fixed standardization technique, with a mean Dice score (DSC in \%) and mean Hausdorff distance (MHD in millimeters) of 86.7\%\,/\,0.444\,mm and 87.0\%\,/\,0.482\,mm respectively, compared to 84.1\%\,/\,0.675\,mm for a conventional Standardization. Surprisingly, our method achieves a performance on par with the learned Pre-processor, obtaining a slightly lower mean Dice score but slightly better mean  Hausdorff distance. This indicates that imposing realism constraints, while also considering the downstream segmentation task, does not impact the segmentation accuracy. We also note that the performance of our joint normalization method, when trained with both iSEG and MRBrainS datasets, is greater than that of the segmentation network when trained and tested independently on these datasets (Table \ref{tab:baseline}). These results demonstrate the benefit of exploiting jointly-normalized datasets to improve the overall performance. Examples of predicted segmentation for iSEG and MRBrainS test images are shown in Fig.~\ref{fig:visualization}. We see that our method yields a segmentation mask close to the ground-truth for the two datasets and for all brain tissue classes. 

\begin{figure}[t!]
\centering
    \includegraphics[width=1.02\linewidth]{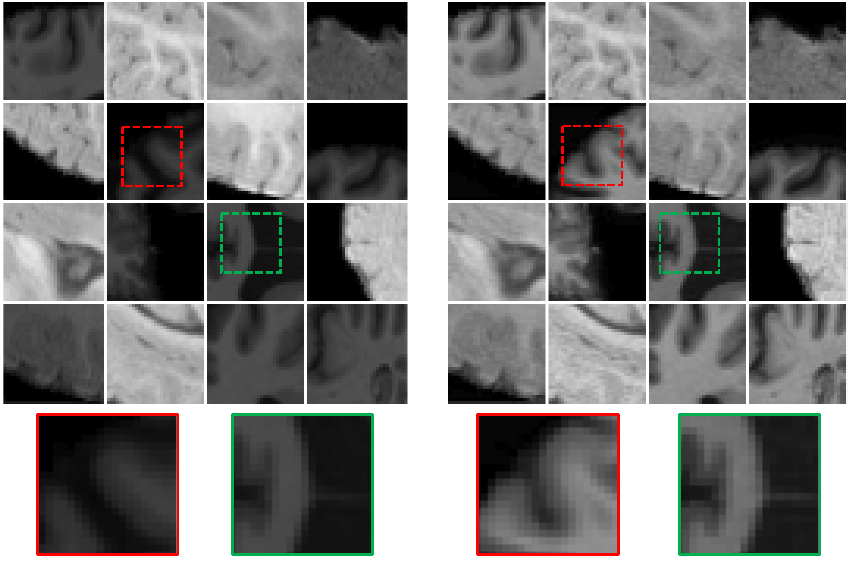}
    
    \vspace*{-2mm} 
    \caption{Mixed iSEG and MRBrainS inputs (\textbf{left}) and the generated images with adversarial normalization (\textbf{right}). Notice the improved homogeneity of intensities in the normalized images and enhanced tissue contrast (see zoomed regions).}
    \label{fig:normalized-patches}
\end{figure}

\begin{figure}[ht!]
    \centering  
    \begin{footnotesize}
    \mbox{    
    \shortstack{
        \includegraphics[width=0.5\linewidth]{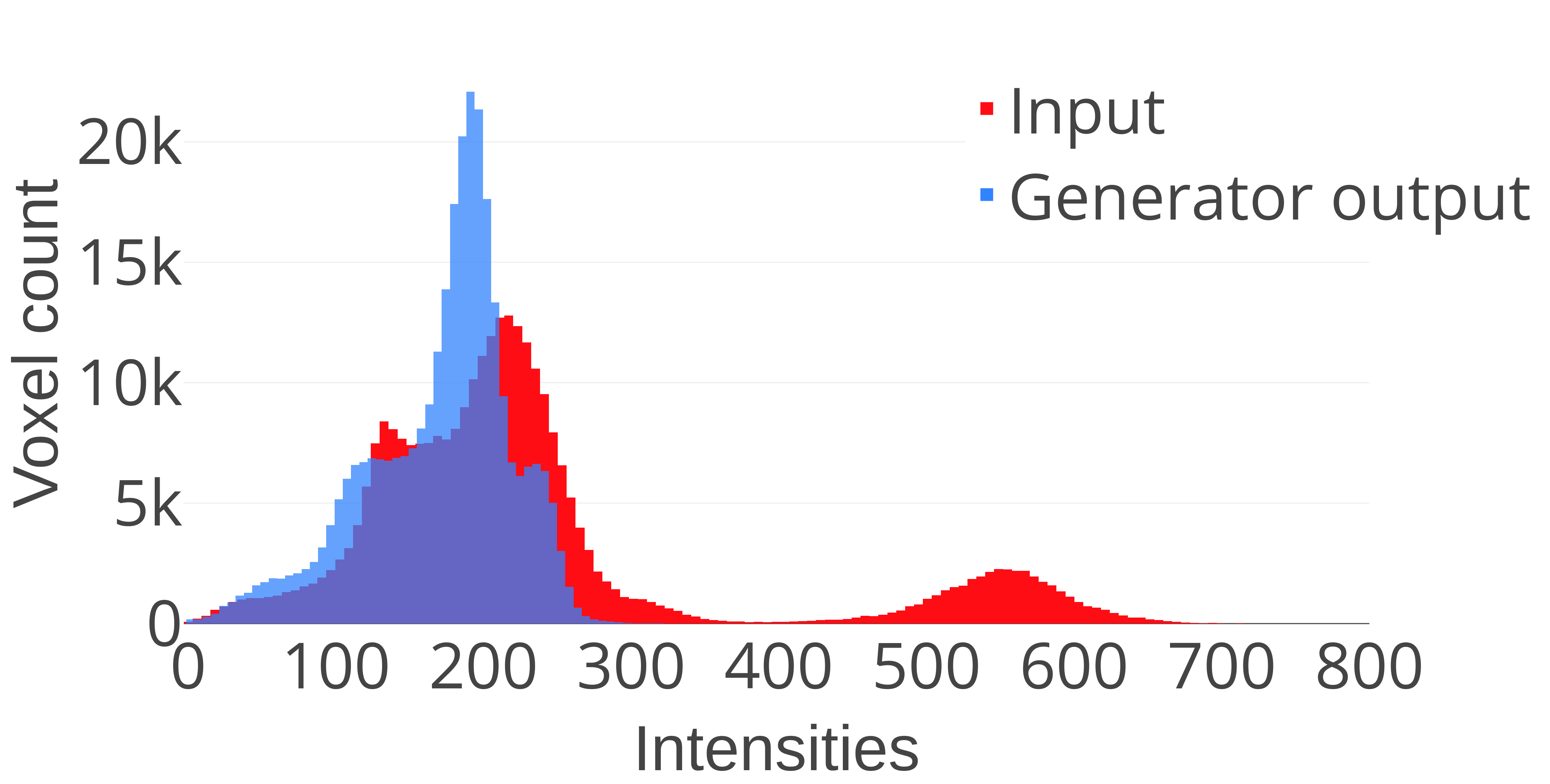} \\[2pt]
    Grey matter (GM)}
    \shortstack{
        \includegraphics[width=0.5\linewidth]{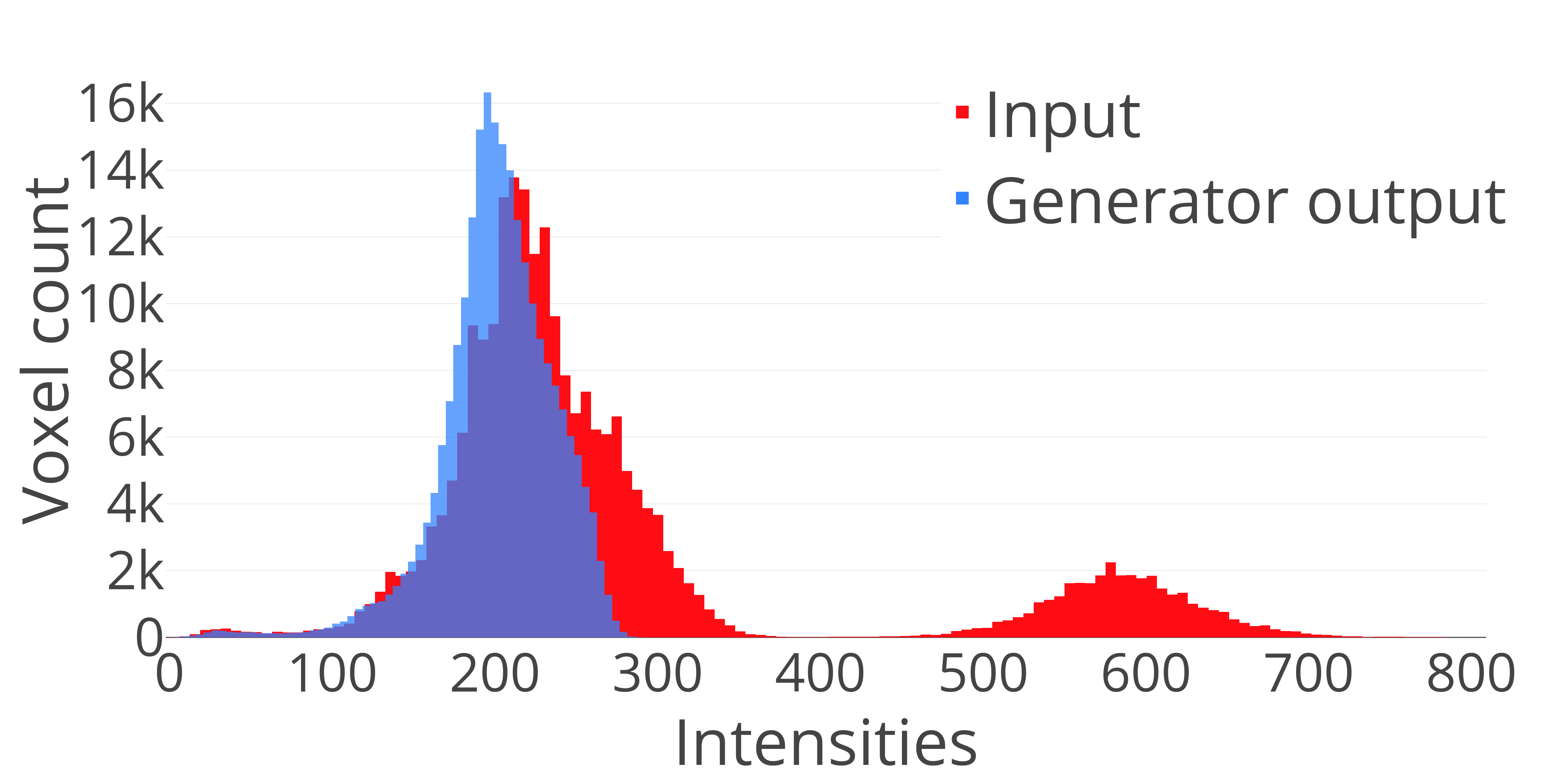} \\[2pt]
    White matter (WM)}
    }
    
    %\vspace{4pt}
    
    \shortstack{    
        \includegraphics[width=0.5\linewidth]{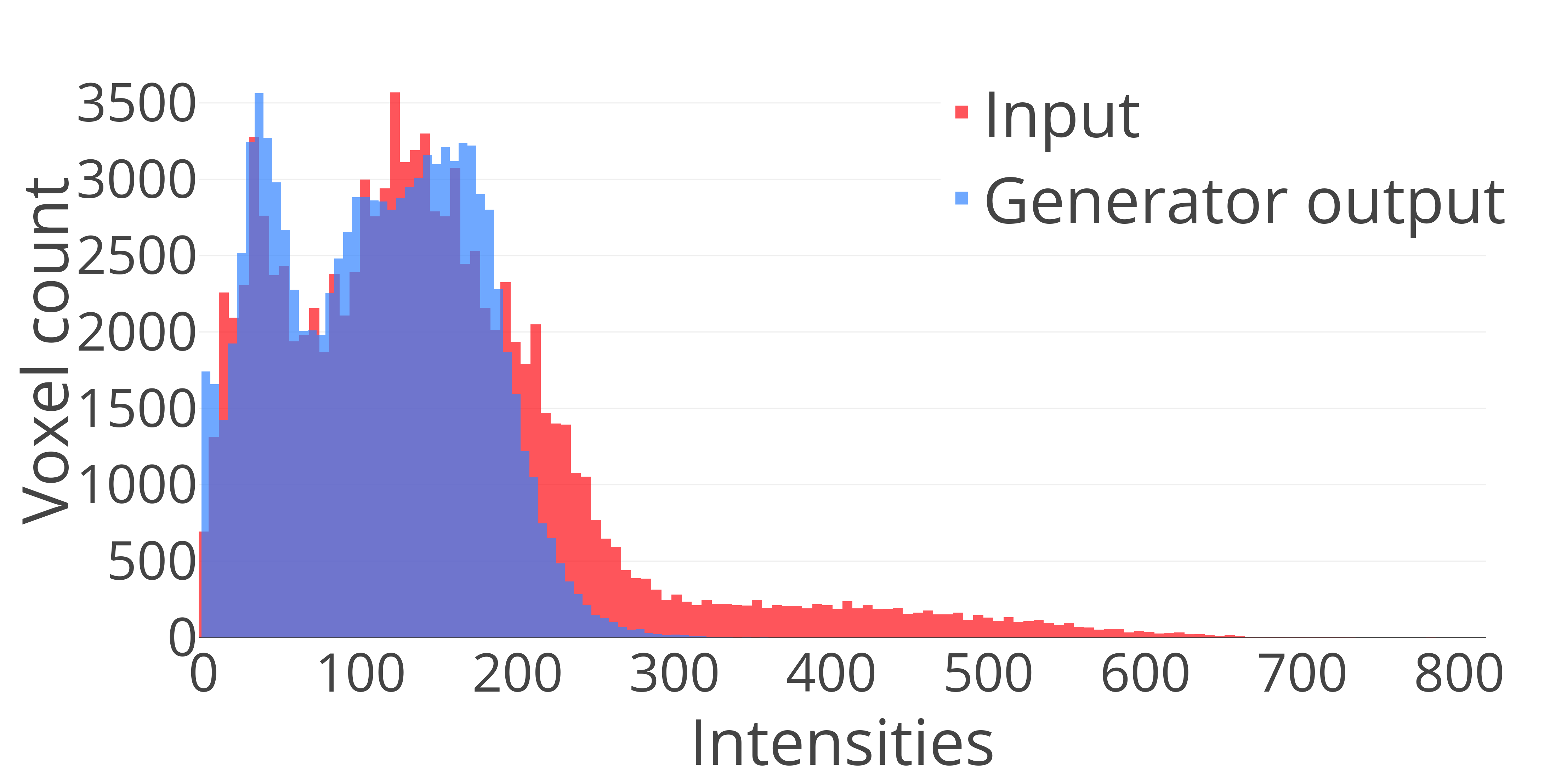} \\[2pt]
    CSF}
    \end{footnotesize}
    
    \vspace*{-2mm} 
    \caption{Histograms of tests images with constraint of realism. One can notice the intensities have been modified and follow a more Gaussian-like curve with overlapping intensities for each classes.}
    \label{fig:histogram_normalized}
\end{figure}

\subsubsection{Normalization Performance}

The advantage of our method in generating realistic images is illustrated in Fig.~\ref{fig:normalized-patches}, where randomly-selected patches from input images and the corresponding output of our generator are shown. Normalized patches present a more uniform intensity distribution, while also preserving the realism and details of the original patches. Our method also exhibits a better contrast enhancement in the generated images, as seen in the zoomed regions of the figure. This is enabled by our task-driven approach which also minimizes the segmentation loss, thereby increasing the contrast along region boundaries.

The normalization effect of our method can be further appreciated from Fig.~\ref{fig:histogram_normalized}, which shows the histogram of intensities for input images and the output of our generator (combined iSEG and MRBrainS images). While input images have a broader spread in distributions with distinct modes, the normalized images have a narrower distribution more centered around a single mode. This helps reducing the intra-class variance and, therefore, increases the segmentation accuracy.  

\begin{table}[t!]
  \centering
  \caption{Jensen-Shannon Divergence (JSD) of input and normalized images from the generator. A lower value corresponds to more similar distributions. Dataset normalization performance is on par with the method in \citet{Drozdzal2018} while still offering visually realistic images.}\label{tab:JSD}
  \begin{footnotesize}
  \renewcommand{\arraystretch}{1.05}
  \begin{tabular}{cccc}
    \toprule
    \B Datasets & \B Input data & \B Pre-processor%~\citep{Drozdzal2018} 
    & \B Ours \\
    \midrule
    iSEG + MRBrainS & \phantom{1}2.0839 & 0.2793 & 0.2788 \\
    iSEG + MRBrainS + ABIDE & 12.2212 & 0.4185 & 0.4180 \\
    \bottomrule
  \end{tabular}
  \end{footnotesize}  
\end{table}

\subsection{Multi-site Evaluation}

In the next experiment, we evaluate our adversarial normalization method in a multi-site scenario involving a third dataset, ABIDE. This large-scale dataset contains images obtained from 17 sites with different acquisition protocols. The segmentation performance of the Standardization technique, the learned Pre-processor approach of~\cite{Drozdzal2018} and our method, when trained with all three datasets, is reported in Table~\ref{tab:comparison} (multi-site setting). Once again, both the Pre-processor and our method achieve a considerable improvement compared to employing a fixed standardization technique. Specifically, our method yields a 2.40\% improvement in mean Dice over this technique. Its performance is also similar to the segmentation-optimized Pre-processor, which does not preserve realism.

Table~\ref{tab:baseline} gives the Jensen-Shannon Divergence (JSD) between the intensity distributions of images from different datasets. Lower values correspond to more similar distributions between input images and the generator output. 
We see an important decrease in JSD when using normalized images. This illustrates the normalization effect of both our method and the learned Pre-processor. Comparing the two approaches, our method leads to slightly smaller JSD values, suggesting a more uniform distribution of intensities across different datasets. Note that this  metric does not evaluate realism of images, which is the main advantage of our method.

\begin{table}[t!]
%\begin{adjustwidth}{-7em}{-5em}
  \centering
  \caption{Dice score of our normalization method when using only T1 images or T1\,+\,T2 images as input. The higher performance for T1\,+\,T2 inputs demonstrates the capability of our method to process multi-modal data.}\label{tab:modality}
  \begin{footnotesize}
  \renewcommand{\arraystretch}{1.05}
  \setlength{\tabcolsep}{3.5pt}
  \begin{tabular}{ccccccccc}
    \toprule
   \multirow[b]{2}{*}{\B Modality} &  \multicolumn{2}{c}{\B CSF} & \multicolumn{2}{c}{\B GM} & \multicolumn{2}{c}{\B WM} & \multicolumn{2}{c}{\B Mean}\\
     \cmidrule(l{6pt}r{6pt}){2-3} \cmidrule(l{6pt}r{6pt}){4-5} \cmidrule(l{6pt}r{6pt}){6-7}
     \cmidrule(l{6pt}r{6pt}){8-9}
     &  \B DSC & \B MHD & \B DSC & \B MHD & \B DSC & \B MHD & \B DSC & \B MHD \\
    \midrule
    T1 only & \B 0.912 & 0.245 &  0.853 & 0.492 &  0.836 & 0.595 & 0.867 &  0.444 \\
    T1\,+\,T2 & 0.910 & \B 0.165 & \B 0.889 & \B 0.431 & \B 0.862 & \B 0.493 & \B 0.887 & \B 0.363 \\
    \bottomrule
  \end{tabular}
  \end{footnotesize}
  %\end{adjustwidth}
\end{table}

\subsection{Multi-modal Testing}

Recent approaches that use multiple image modalities can increase the accuracy of learned tasks~\citep{Dolz2019}. For instance, T1 images typically provide a higher contrast between gray and white matter tissues, while T2 images offer a better contrast between brain tissue and CSF. Combining both T1 and T2 is, therefore, expected to improve brain segmentation. We assess the improvement in segmentation accuracy of our normalization method when used with multi-modal images. 

Table~\ref{tab:modality} compares the Dice scores obtained by our method when using T1 images only as input or both pre-aligned T1 and T2 images as inputs. The accuracy is measured as higher when employing the two MRI sequences, with a mean Dice score improvement of 0.020 and an average  Hausdorff Distance improvement of 0.081\,mm.

\begin{table}[t!]
  \centering
   \caption{Pearson correlation coefficient ($\rho$) between voxel intensity and y-axis position in an axial slice of an MRBrainS and iSEG test subject before and after normalization. A higher correlation implies a stronger bias field degradation. Note that the proposed method reduces correlation in all cases. %\textbf{\color{red}HL: Perhaps bold the best score?}
  }\label{tab:denoising}
  \begin{footnotesize}
  \renewcommand{\arraystretch}{1.05}
  \begin{tabular}{ccccccc}
    \toprule
    %\multirow[b]{3}{*}{{\B $\alpha$}} &
    %\multicolumn{4}{c}{\B %Pearson Correlation Coefficient ($\rho$)} \\
    %\cmidrule(l{6pt}r{6pt}){2-5}
    \multirow[b]{2}{*}{{\B $\boldsymbol{\alpha}$}} & \multicolumn{2}{c}{\B MRBrainS} & \multicolumn{2}{c}{\B iSEG}\\
    \cmidrule(l{6pt}r{6pt}){2-3}
    \cmidrule(l{6pt}r{6pt}){4-5}
    & \B Degraded &  \B Normalized & \B Degraded & \B Normalized\\    
    \midrule 

    Original & $-$0.0199 & -- & 0.0271 & -- \\   
    %0.1 & \phantom{$-$}0.0158 & $-$0.0193 & 0.0980 & 0.0433\\   
    0.3 & \phantom{$-$}0.0982 & \phantom{$-$}0.0534  & 0.2560 & 0.1192\\    
    0.5 & \phantom{$-$}0.1963 & \phantom{$-$}0.1422  & 0.4264 & 0.2430\\    
    0.7 & \phantom{$-$}0.3103 & \phantom{$-$}0.2393  & 0.5914 & 0.4599\\    
    0.9 & \phantom{$-$}0.4368 & \phantom{$-$}0.3649  & 0.7316 & 0.5928\\    
    \bottomrule
  \end{tabular}
  \end{footnotesize}
\end{table}

\begin{figure}[!t]
\centering    
    \includegraphics[width=.75\linewidth]{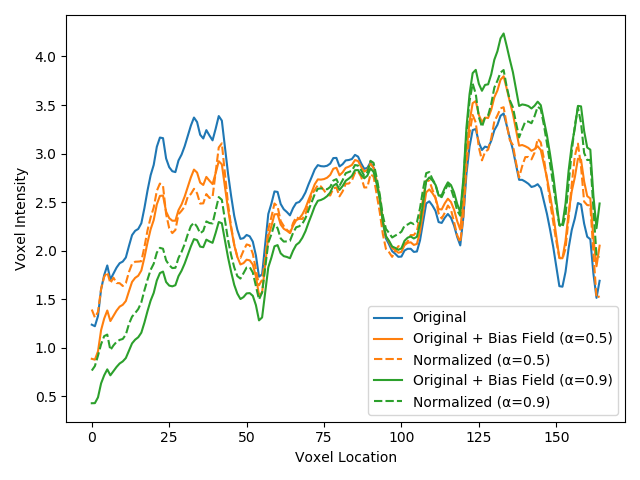}\\
    \includegraphics[width=.75\linewidth]{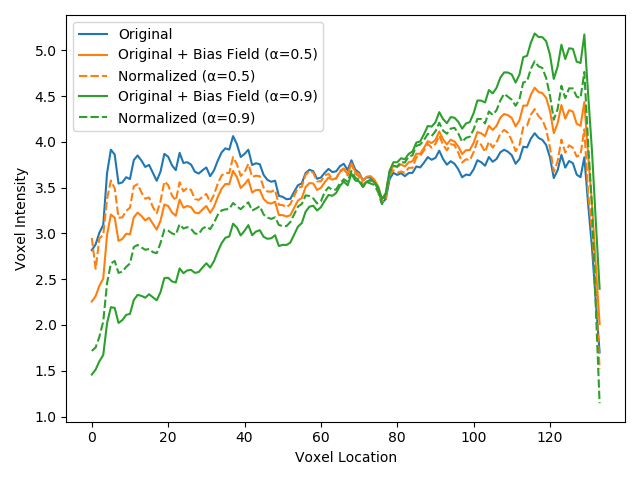}
    \vspace*{-2mm}
    \caption{Mean voxel intensity in an axial slice of an MRBrainS (left) and iSEG (right) test subject across the brain for bias field strength of $\alpha$\,=\,0.5 and $\alpha$\,=\,0.9. By bringing intensity values closer to those of the original image, our normalization method helps correct the bias field in degraded images.
    }
    \label{fig:bias_field_graph}
\end{figure}

\subsection{Robustness to Image Degradation}

Our task-driven normalization method also demonstrates the ability to enhance images with non-homogeneous intensity. To evaluate this, we trained the method with 50\% of input images augmented with a random bias field as described in Section~\ref{sec:pre-processing}. Since the discriminator must discern the generator output for these degraded images from the real, non-degraded images, it encourages the generator to remove the bias field while also preserving realism. 

We measure our method's ability to correct the bias field by computing the Pearson correlation between the intensity and position of voxels along the field's direction (i.e., y-axis). Since we used a linear bias field, a higher correlation corresponds to a stronger degradation of the image. Table \ref{tab:denoising} gives the mean correlation for test images of the MRBrainS and iSEG datasets, the same images degraded with a bias field of increasing strength $\alpha \in \{0.3,\, 0.5,\, 0.7,\, 0.9\}$, and the output of our generator for these images. The mean intensity as function of the y-axis position, for $\alpha$\,=\,0.5 and $\alpha$\,=\,0.9, is shown in Fig.~\ref{fig:bias_field_graph}. We observe that, for both datasets, intensity is noticeably less correlated to the y-axis position in normalized images than in degraded images, illustrating our method's ability to correct inhomogeneity. The benefit of our method is particularly important for stronger bias fields (i.e., $\alpha$\,=\,0.7 and $\alpha$\,=\,0.7). An example of a test image with bias field of strength $\alpha$\,=\,0.5 and corresponding normalized output of the generator is provided in Figure~\ref{fig:example_degraded}. One can see that intensities in the normalized image are more uniform.

\begin{table}[t!]
  \centering
  \caption{Mean Dice score for different values of hyper-parameter $\lambda$. A lower $\lambda$ emphasizes segmentation performance while a higher value gives more importance to the realism of the generated image. 
  }\label{tab:lambdaComparison}
  \begin{footnotesize}
  \renewcommand{\arraystretch}{1.05}
  \setlength{\tabcolsep}{8pt}
  \begin{tabular}{cccc}
    \toprule
    \multirow[b]{2}{*}{\B Lambda\,($\lambda$)} & \multirow[b]{2}{*}{\B Mean DSC} & \multicolumn{2}{c}{\B Discr. accuracy\,(\%)}\\
    \cmidrule(l{6pt}r{6pt}){3-4}
    & & \B Train & \B Test \\
    \midrule 
   % 0.0001 & 0.870 & 89.65 & 29.92 \\
%    0.001 & 0.868 & &  \\
 %   0.01 & 0.872 & & \\
    0.1 & 0.875 & 98.95 & 34.49  \\
    1.5 & 0.866 & 44.10 & 32.58 \\
    5.0 & 0.851 & 44.07 & 28.58 \\
    \bottomrule
  \end{tabular}
  \end{footnotesize}
  
  %\vspace*{-2mm}  
\end{table}

\begin{figure*}[t!]
    \centering
    \setlength{\tabcolsep}{5pt}
    \begin{tabular}{ccc}
    \confmat{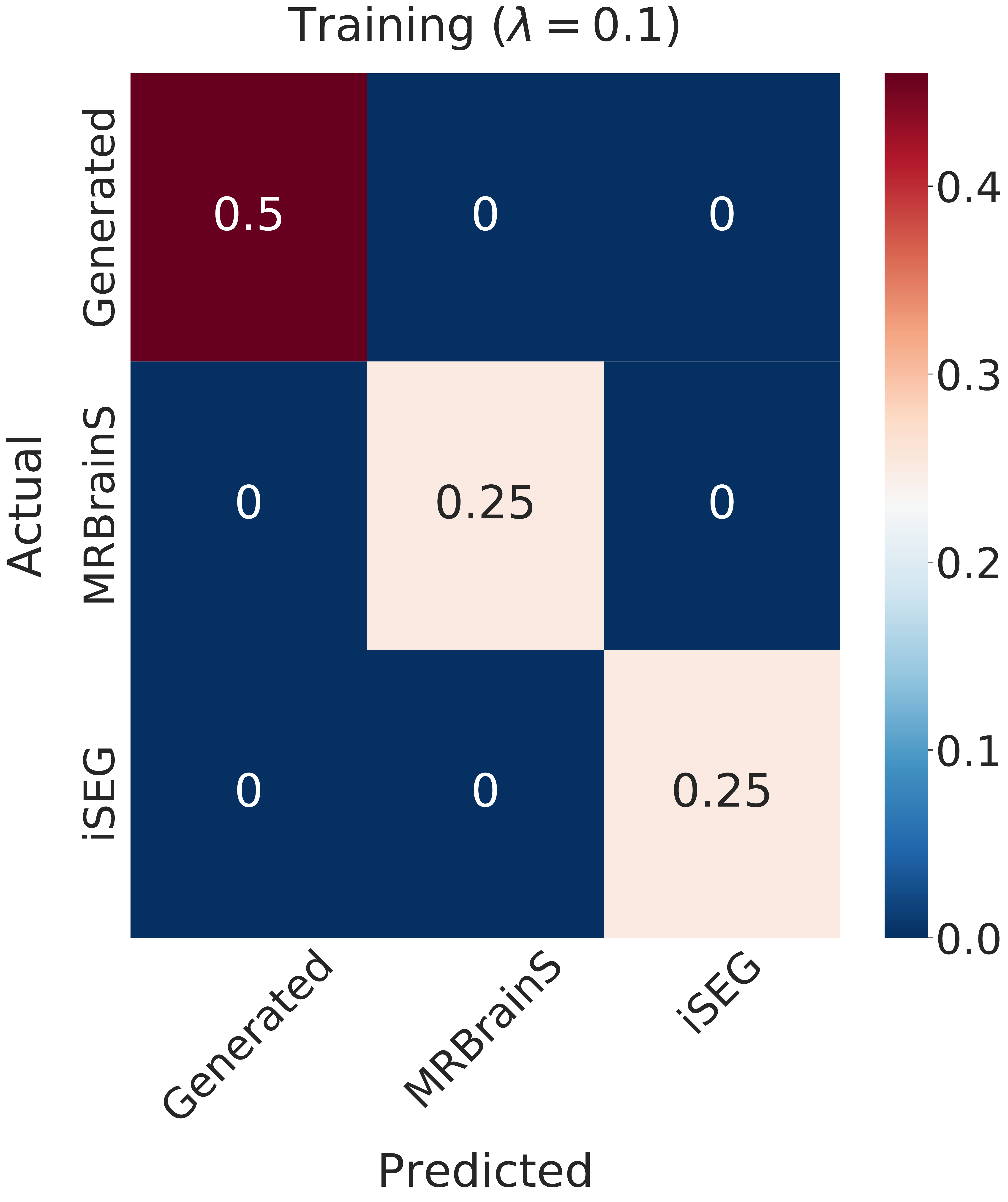} & \confmat{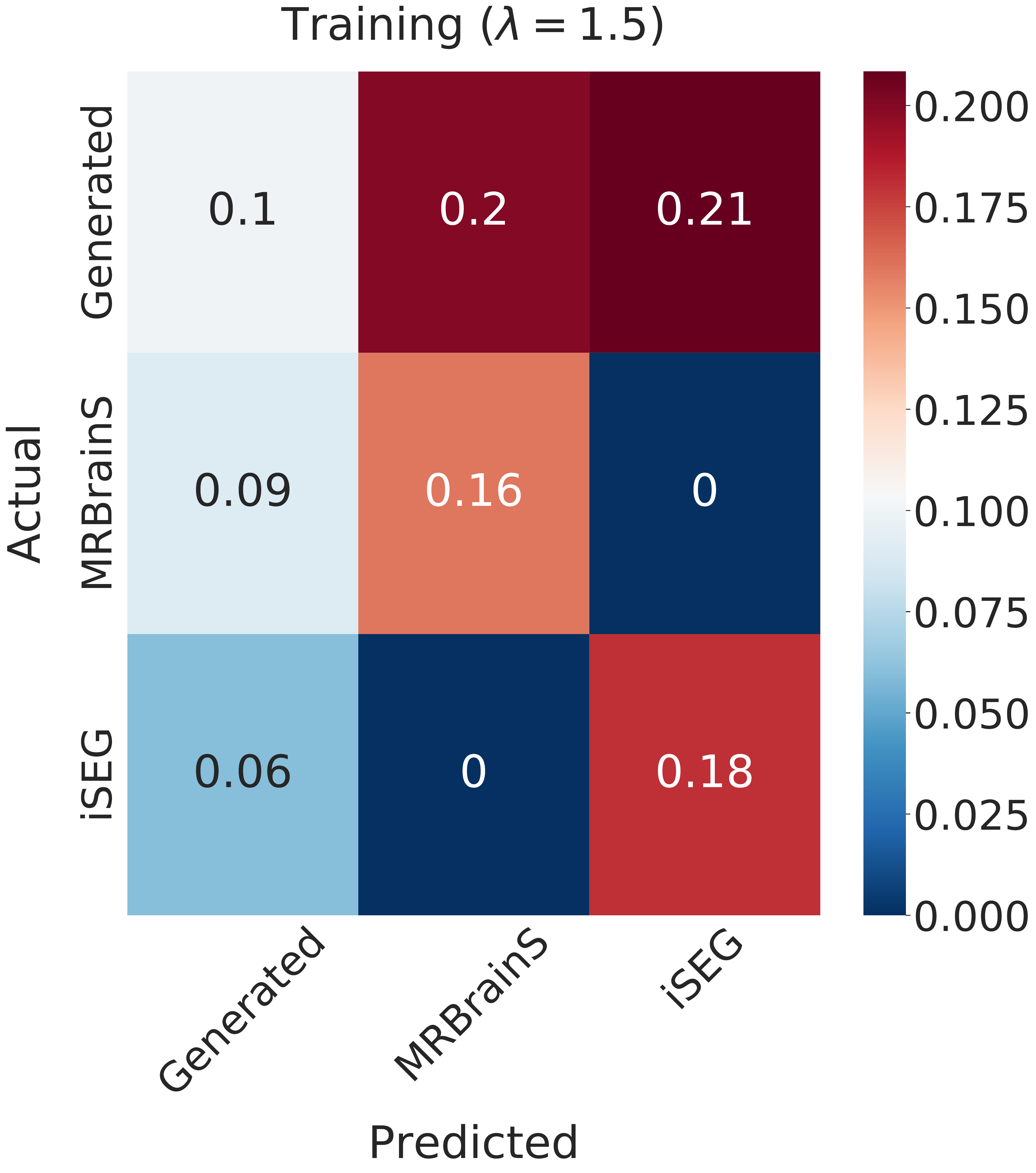} & 
    \confmat{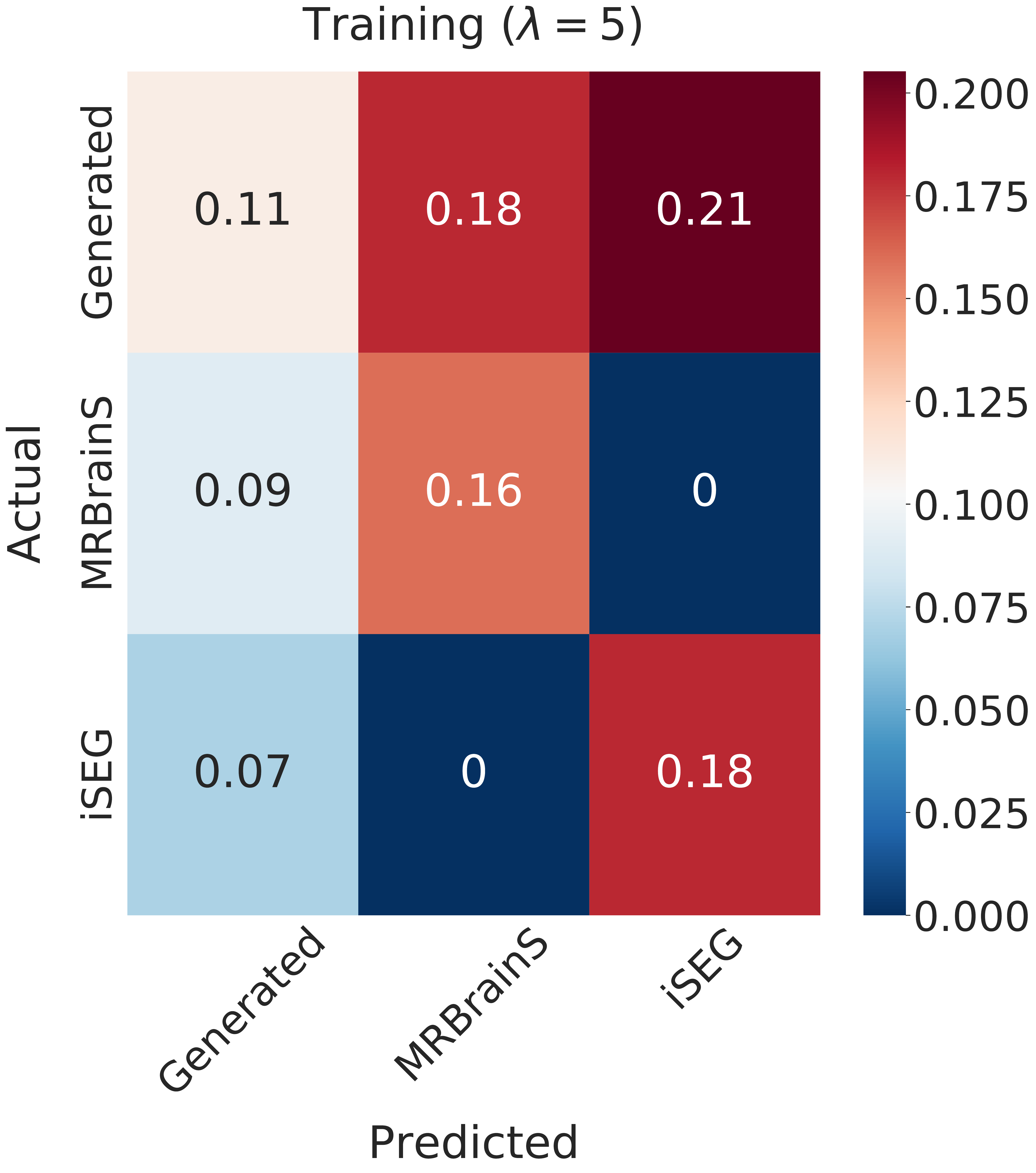} \\[1mm]
    \confmat{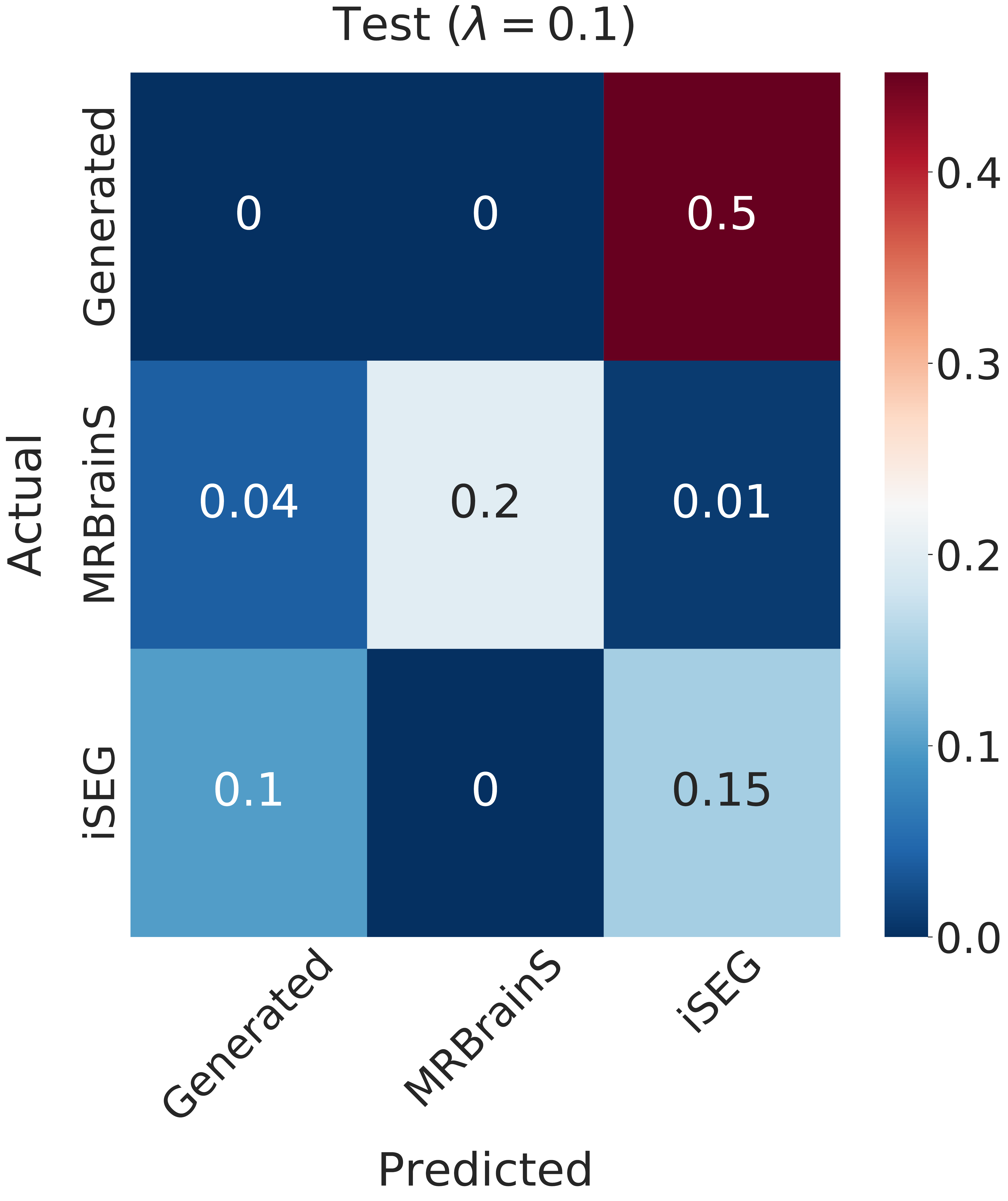} & \confmat{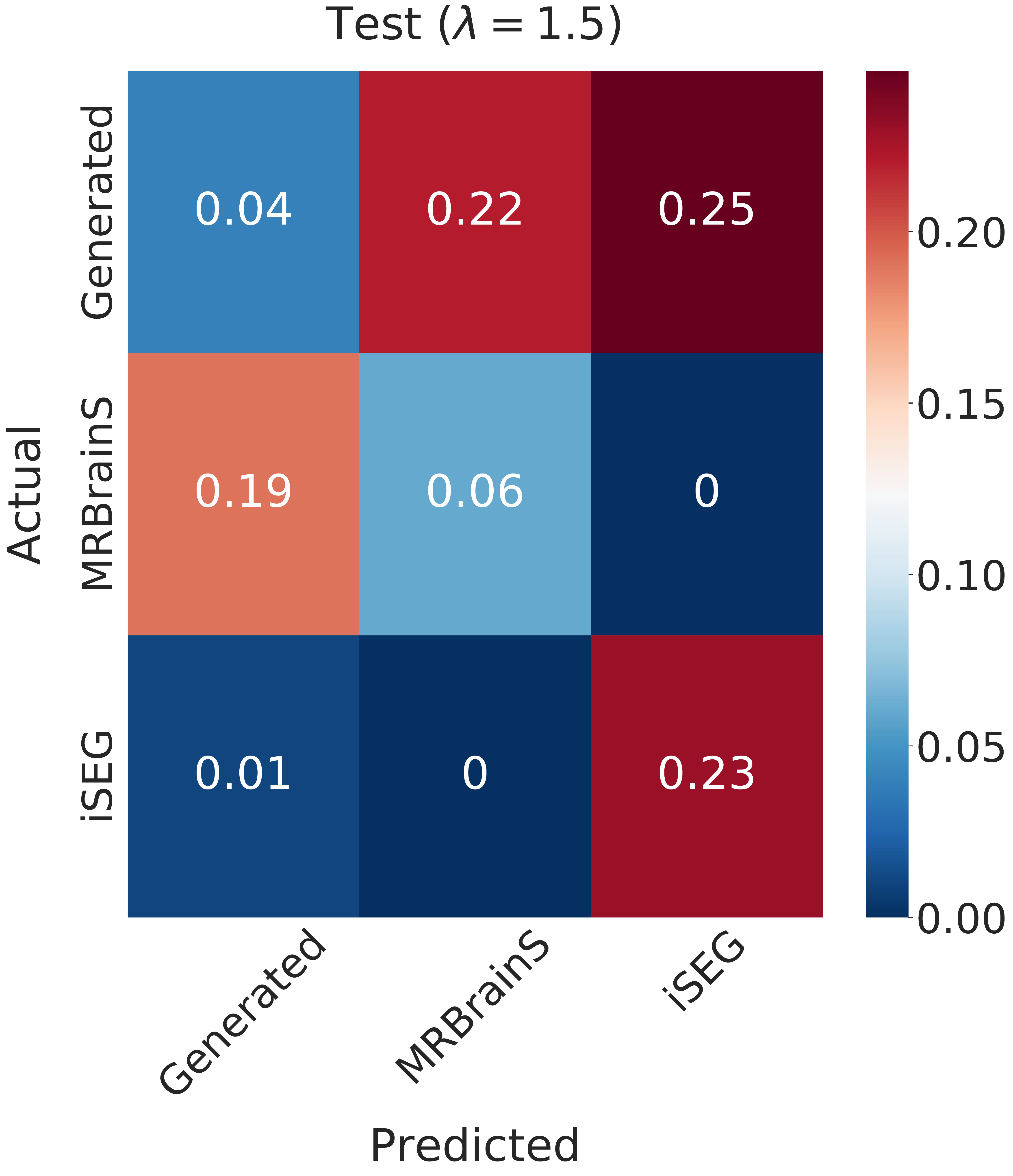} & 
    \confmat{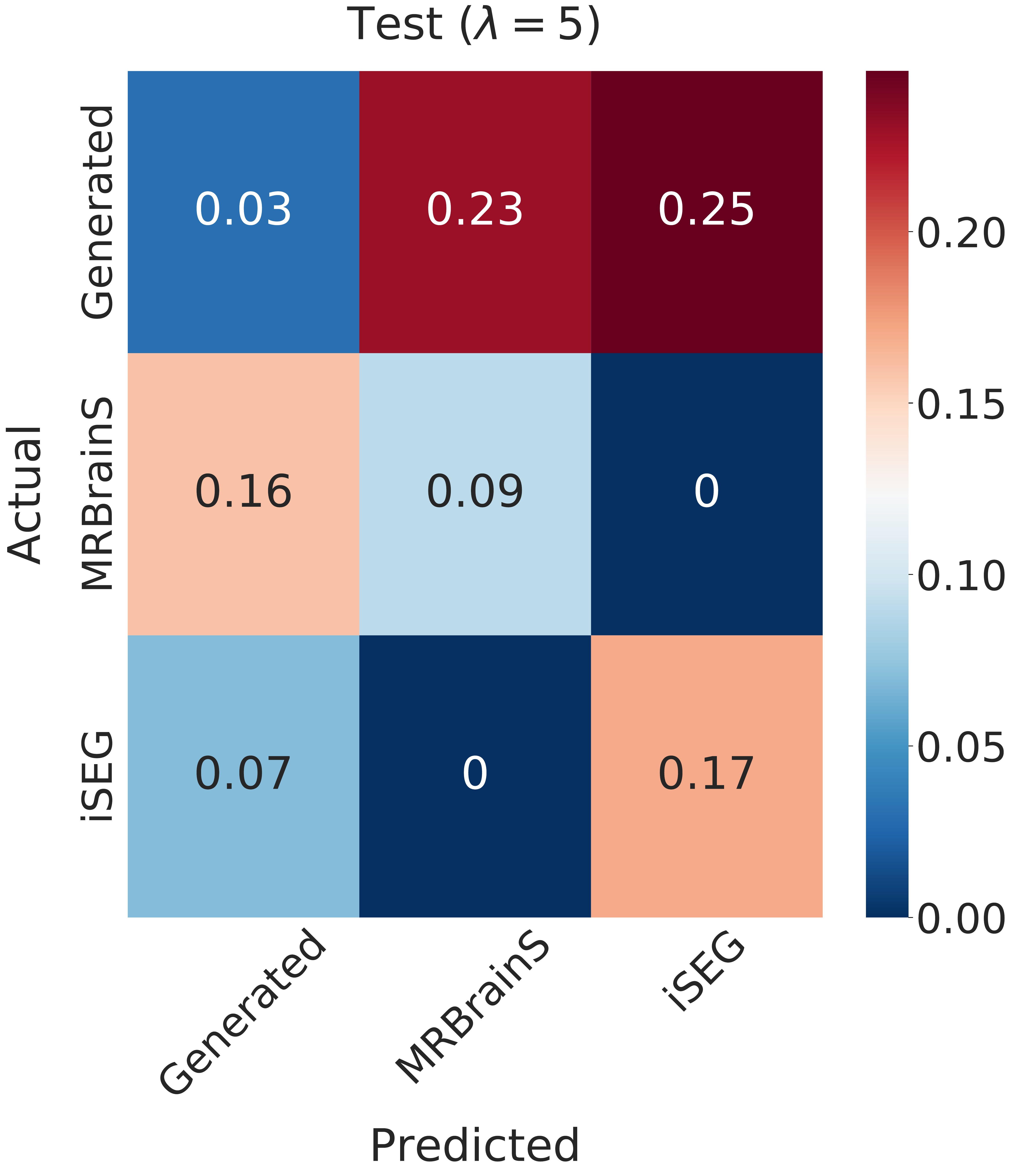}
    \end{tabular}
    
    \vspace*{-2mm}
    \caption{Normalized confusion matrix of the domain classifier (discriminator) for different values of hyper-parameter $\lambda$. \textbf{Top row}: training examples. \textbf{Bottom row}: test examples. For $\lambda$\,=\,1 generated training images are easily differentiated from real ones. In contrast, when using $\lambda$\,$\geq$\,1.5, the generator confuses generated images for real ones, with a uniform distribution between the MRBrainS and iSEG classes.
    }
    \label{fig:confusion_matrix}
\end{figure*}

\subsection{Impact of Hyper-parameter Lambda}

As defined in Eq.~(\ref{eq:total_loss}), hyper-parameter $\lambda$ has a direct impact on the level of realism in generated images. A lower value emphasizes segmentation accuracy, while a higher value prioritizes the generation of more realistic, domain-invariant images. In this last experiment, we analyse the impact of this crucial hyper-parameter on image segmentation and normalization. 

Table~\ref{tab:lambdaComparison} gives the mean Dice scores and discriminator accuracy for our training and test samples with $\lambda \in \{0.1,\, 1.5,\, 5.0\}$. As expected, the segmentation accuracy decreases with higher $\lambda$ values, since the model focuses more on generating realistic images and less on obtaining a precise segmentation. Hence, we observe a 2.4\% drop in mean Dice scores when increasing from $\lambda$\,=\,0.1 to $\lambda$\,=\,5.0. The realism of the generated images can be estimated with the discriminator accuracy. A higher value indicates that the generated images can be more easily differentiated from real ones. For a small $\lambda$ value of 0.1, the discriminator can classify almost perfectly all our training examples, indicating that images produced by the generator are indeed very different from real images. As $\lambda$ is increased, generated images become more similar to real ones, resulting in a lower discriminator accuracy both in training and testing samples. 

The behavior of our discriminator for the $\lambda$ values is further analyzed in Fig.~\ref{fig:confusion_matrix} which shows the normalized confusion matrix\footnote{The matrix is normalized by dividing each value by the total number of samples.} for training and testing samples. The normalized training images are correctly identified as generated with $\lambda$\,=\,0.1, but uniformly predicted as MRBrainS or iSEG with $\lambda$\,=\,1.5 or $\lambda$\,=\,5.0. This corresponds to the scenario expected from Theorem \ref{th1}, where generated images from different datasets follow the same distribution and the discriminator predicts domain classes uniformly. We also note that the discriminator overfits the training data with $\lambda$\,=\,0.1, resulting in a poor classification of the generated samples during testing.

\section{Conclusions}\label{sec:conclusion}

This paper presents a realistic task-and-data-driven normalization method that improves the segmentation of images by exploiting multiple datasets simultaneously. Our method leverages an adversarial learning strategy that involves three networks: a generator which normalizes input images while preserving their realism, a task network that predicts an accurate segmentation from normalized images, and a discriminator which classifies the domain of these images. Unlike traditional adversarial approaches for image synthesis or domain adaptation, where a discriminator distinguishes real images from fake ones or predicts the domain of images, our discriminator is trained in a ($K$+1)-class classification problem with $K$ domain classes corresponding to the originating dataset (or site) of real images, and an additional class corresponding to the generated normalized images. By maximizing the discriminator loss and simultaneously minimizing the segmentation loss during training, the generator consequently learns to produce images that are both harmonized and realistic across all datasets, while still optimizing for segmentation. Compared to the recent data harmonization techniques~\citep{modanwal2020mri}, our method has less hyper-parameters to tune and can more easily adapt to the addition of new datasets or image modalities. 

The advantage of our method has been demonstrated in a comprehensive set of experiments involving three largely different brain MRI datasets: iSEG, MRBrainS and ABIDE. In an experiment with iSEG and MRBrainS images, we first established that a standard segmentation network performs poorly when trained and tested across different datasets (Table~\ref{tab:baseline}). In contrast, our adversarial normalization method achieves a consistently higher accuracy using images with dissimilar intensity distributions (Table~\ref{tab:comparison} and Fig.~\ref{fig:visualization}). Moreover, our normalization network also provides clinically interpretable images when compared to state-of-the-art approaches~\citep{Drozdzal2018} (Fig.~\ref{fig:normalized-patches}). 

Our experiments also showed our normalization strategy to yield good performance on data from more than two sites or with multiple image modalities. While trained on all three datasets, our method achieved a higher mean Dice score of 0.890 compared to 0.867 when employing only iSEG and MRBrainS (Table~\ref{tab:comparison}). In addition, it considerably reduced the variability of intensities in normalized images from the different datasets with a JSD of 0.418 compared to 12.221 in the case of unnormalized images (Table~\ref{tab:JSD}). Furthermore, when employing an additional image modality as input (T2-weighted MRI), our method also obtained a better segmentation accuracy, with a mean Dice improvement of 0.02 with respect to using only T1-weighted images (Table \ref{tab:modality}).

We further evaluated our method by analyzing its robustness to image degradation and its sensitivity to hyper-parameters. In addition to harmonizing images from different sites, our adversarial normalization method could also remove intensity inhomogeneity without requiring additional processing (Table \ref{tab:denoising}). This could help achieve a much faster and more reliable analysis of large-scale brain MRI datasets compared to traditional processing pipelines  (Table~\ref{tab:lambdaComparison} and Fig.~\ref{fig:confusion_matrix}). Training our model with different values of hyper-parameter $\lambda$ enabled us to study the trade-off between segmentation accuracy and normalized image realism. Best results were found with $\lambda$\,=\,1.5.    

Our task-driven adversarial normalization approach unlocks the training of deep learning models with data from multiple sites by improving both realism and accuracy of normalized images across datasets. A potential technical limitation of the proposed method is the need to process 3D images in smaller sub-regions due to the current limitation in GPU memory. Although we obtained spatially-smooth generated images and segmentation maps, the overall result may be sub-optimal since the global context and intensity distribution of images is not fully considered when working with local patches. In future work, we plan to tackle this problem by incorporating global image statistics in the loss function, and by exploring a 2.5D approach~\citep{xue2020multi} where slices from different view planes are processed simultaneously. Moreover, while the discriminator architecture employed in our model generally led to good results, it can sometimes suffer from vanishing gradient or mode collapse. Our future work will also investigate adversarial models such as the Wasserstein GAN \citep{Arjovsky2017} which are less prone to these problems. Finally, we aim to demonstrate our method in a broader set of applications, including the segmentation of brain lesions where preserving fine regions during normalization is critical. 

The source code of this article is publicly available on \href{https://github.com/sami-ets/DeepNormalize}{Github}\footnote{\url{https://github.com/pldelisle/deepNormalize}}.

\section*{Acknowledgments}
This work was partially supported by the Canada Research Chair on Shape Analysis in Medical Imaging, the Research Council of Canada (NSERC), the Fonds de Recherche du Quebec (FQRNT), and ETS Montreal. The authors also thank NVIDIA for the donation of a GPU. 

\appendix

\section{Proof of Theorem 1}
\label{sec:proof}

\begin{thm}\label{th1}
Let $p_r(\xx \, | \, z)$ and $p_g(\xx \, | \, z)$ be the probabilities that $\xx$ is a real or a generated image, respectively, from source dataset $z$. The minimax optimization problem of Eq. (\ref{eq:total_loss}) without the segmentation term corresponds to minimizing the divergence between $p_g(\xx \, | \, z)$ for each $z$ and the mean distribution of real images $\overline{p}_r(\xx) = \frac{1}{k}\sum_{z=1}^K\, p_r(\xx \, | \, z)$.
\end{thm}

\begin{pf} If we ignore the segmentation loss $\lossSeg$, the optimization problem is given by
\begin{equation}\label{eq:minimax_opt}
\begin{aligned}
\min_{G} \, \max_{D} \ \loss(G,D) & \ = \ \expect_{\xx,z} \Big[\log D_z(\xx) \, + \, \log D_{K+1}(G(\xx))\Big]\\
& \ = \ \expect_{\xx,z} \Big[\log D_z(\xx) \ + \ \log \Big(1 - \sum_{z'=1}^K D_{z'}(G(\xx))\Big)\Big]
\end{aligned}
\end{equation}
Suppose generator $G$ is fixed, the optimal discriminator $D^*$ is found by minimizing
\begin{equation}
\loss_G(D) \, = \, -\sum_{z=1}^K p(z) \int_{\xx} \Big[p_r(\xx \, | \, z)\log D_z(\xx) \, + \, p_g(\xx \, | \, z) \log \Big(1\!-\! \sum_{z'=1}^K D_{z'}(\xx)\Big)\Big] \mathrm{d}\xx.
\end{equation}
We obtain the optimum value for each $\xx$ by deriving this function with respect to $D_z(\xx)$
\begin{equation}
\frac{\partial\loss_D}{\partial D_z(\xx)} \ = \ 
-\frac{p(z)\, p_r(\xx \, | \, z)}{D_z(\xx)} \ + \ \frac{p(z)\, p_g(\xx \, | \, z)}{1- \sum_{z'=1}^K D_{z'}(\xx)}.
\end{equation}
Setting this to zero yields
\begin{equation}
\frac{D^*_z(\xx)}{1-\sum_{z'=1}^K D^*_{z'}(\xx)} \ = \ \frac{p_r(\xx \, | \, z)}{p_g(\xx \, | \, z)}.
\end{equation}
Summing both sides of the equation over $z$, and using $D_{K+1}(\xx) = 1 - \sum_z D_z(\xx)$, we then get
\begin{equation}
\frac{\sum_{z=1}^K D^*_z(\xx)}{1-\sum_{z'=1}^K D^*_{z'}(\xx)} \ = \
\sum_{z=1}^K \frac{p_r(\xx \, | \, z)}{p_g(\xx \, | \, z)} \ = \ 
\frac{1-D^*_{K+1}(\xx)}{D^*_{K+1}(\xx)}
\end{equation}
and therefore
\begin{align}
D^*_{z}(\xx) & \ = \ \frac{\frac{p_r(\xx \, | \, z)}{p_g(\xx \, | \, z)}}{1 \, + \, \sum_{z'=1}^K \frac{p_r(\xx \, | \, z')}{p_g(\xx \, | \, z')}}, \quad z=1,\,\ldots,\,K; \\
D^*_{K+1}(\xx) & \ = \ \frac{1}{1 \, + \, \sum_{z'=1}^K \frac{p_r(\xx \, | \, z')}{p_g(\xx \, | \, z')}}\label{eq:dfake_opt}
\end{align}
Next, we use this result to find the optimal generator. Toward this goal, we plug (\ref{eq:dfake_opt}) into the loss of Eq. (\ref{eq:minimax_opt}) and minimize
\begin{equation}
\begin{aligned}\label{eq:opt_gen}
\loss_{D^*}(G) & \ = \ \sum_{z=1}^K\, p(z) \int_{\xx} p_g(\xx \, | \, z) \,\, \log \, D^*_{K+1}(\xx) \,\, \mathrm{d}\xx\\
& \ = \ \sum_{z=1}^K\, p(z) \int_{\xx} p_g(\xx \, | \, z) \, \log \left(\frac{1}{1 \, + \, \sum_{z'=1}^K \frac{p_r(\xx \, | \, z')}{p_g(\xx \, | \, z')}} \right) \mathrm{d}\xx\\
& \ = \ \sum_{z=1}^K\, p(z) \int_{\xx} p_g(\xx \, | \, z) \, \log \left(\frac{p_g(\xx \, | \, z)}{p_g(\xx \, | \, z) \, + \, \sum_{z'=1}^K \frac{p_g(\xx \, | \, z)}{p_g(\xx \, | \, z')}\, p_r(\xx \, | \, z')} \right) \mathrm{d}\xx
\end{aligned}
\end{equation}
Let $q(\xx\,|\,z)$ the probability distribution defined as
\begin{equation}\label{eq:prob_q}
q(\xx\,|\,z) \ = \ \frac{1}{\mathcal{Z}(z)} \Big[p_g(\xx \, | \, z) \, + \, \sum_{z'=1}^K \frac{p_g(\xx \, | \, z)}{p_g(\xx \, | \, z')}\, p_r(\xx \, | \, z')\Big],
\end{equation}
where $\mathcal{Z}(z)$ is a normalization constant. Suppose that the generator produces images similar to the input, i.e. $p_g(\xx \, | \, z') \approx p_r(\xx \, | \, z')$, this constant can be estimated as
\begin{equation}
\mathcal{Z}(z) \ = \ \int_\xx p_g(\xx \, | \, z)\, \mathrm{d}\xx \ + \ \sum_{z'=1}^K \int_x \frac{p_g(\xx \, | \, z)}{p_g(\xx \, | \, z')}\, p_r(\xx \, | \, z')\, \mathrm{d}\xx \ \approx \ K\!\!+\!1.
\end{equation}
Hence, we can rewrite the generator loss in (\ref{eq:opt_gen}) as
\begin{equation}
\begin{aligned}
\loss_{D^*}(G) &  \ = \ \sum_{z=1}^K\, p(z) \int_{\xx} p_g(\xx \, | \, z) \, \log \left(\frac{p_g(\xx \, | \, z)}{(K\!\!+\!1)\,\,q(\xx\,|\,z)} \right) \mathrm{d}\xx\\
&  \ = \ \sum_{z=1}^K\, p(z) \, D_{\mr{KL}}\big(p_g(\cdot \, | \, z) \: || \: q(\cdot\,|\,z)\big) \ - \ \log(K\!\!+\!1),
\end{aligned}
\end{equation}
where $D_{\mr{KL}}(p \: || \: q) \geq 0$ is the KL divergence. Assuming that $p(z)$ is uniform, the optimal generator $G^*$ is therefore such that $p_g(\cdot \, | \, z) = q(\cdot \, | \, z)$, for $z=1,\,\ldots,\,K$. Considering Eq. (\ref{eq:prob_q}), this can only be achieved if $p_g(\cdot \, | \, z) = p_g(\cdot \, | \, z')$, $\forall z,z'$. This in turn gives
\begin{equation}
q(\xx\,|\,z) \ = \ \frac{1}{K\!\!+\!1}\Big[p_g(\xx \, | \, z) \ + \ \sum_{z'=1}^K p_r(\xx \, | \, z')\Big].
\end{equation}
Using the fact that $p_g(\xx \, | \, z) = q(\xx\,|\,z)$, we finally obtain
\begin{equation}
p_g(\xx \, | \, z) \ = \ \frac{1}{K} \sum_{z'=1}^K p_r(\xx \, | \, z') \ = \ \overline{p}_r(\xx).
\end{equation}
Hence, $G^*$ will produce outputs from the average of real image distributions over the different data sources.
%% END PROOF
\end{pf}

%%Harvard
\bibliographystyle{model2-names.bst}\biboptions{authoryear}
\bibliography{sample}

%\section*{Supplementary Material}

%Supplementary material that may be helpful in the review process should
%be prepared and provided as a separate electronic file. That file can
%then be transformed into PDF format and submitted along with the
%manuscript and graphic files to the appropriate editorial office.

\end{document}